\documentclass[sigconf]{acmart}

\usepackage{caption}
\usepackage{subcaption}
\usepackage{soul}

\AtBeginDocument{%
  \providecommand\BibTeX{{%
    \normalfont B\kern-0.5em{\scshape i\kern-0.25em b}\kern-0.8em\TeX}}}

\newcommand\blfootnote[1]{%
  \begingroup
  \renewcommand\thefootnote{}\footnote{#1}%
  \addtocounter{footnote}{-1}%
  \endgroup
}

\begin{document}
\fancyhead{}

% remove footnote with conference information in first column 
\renewcommand\footnotetextcopyrightpermission[1]{} 

%%%%%%% To remove copyright notice %%%%%%%
\makeatletter
\def\@copyrightspace{\relax}
\makeatother

%%
%% The "title" command has an optional parameter,
%% allowing the author to define a "short title" to be used in page headers.
% \title{CAVE - Concerns About Vaccines with Explanations}

%\title{A dataset for explainable classification and summarization of concerns towards COVID vaccines}

\title{CAVES: A Dataset to facilitate Explainable Classification and Summarization of Concerns towards COVID Vaccines*}

%%
%% The "author" command and its associated commands are used to define
%% the authors and their affiliations.
%% Of note is the shared affiliation of the first two authors, and the
%% "authornote" and "authornotemark" commands
%% used to denote shared contribution to the research.
% \author{Ben Trovato}
% \authornote{Both authors contributed equally to this research.}
% \email{trovato@corporation.com}
% \orcid{1234-5678-9012}
% \author{G.K.M. Tobin}
% \authornotemark[1]
% \affiliation{%
%   \institution{Institute for Clarity in Documentation}
%   \streetaddress{P.O. Box 1212}
%   \city{Dublin}
%   \state{Ohio}
%   \country{USA}
%   \postcode{43017-6221}
% }

\author{Soham Poddar}
\affiliation{%
 \institution{Indian Institute of Technology}
 \city{Kharagpur}
 \country{India}}

\author{Azlaan Mustafa Samad}
\affiliation{%
 \institution{Indian Institute of Technology}
 \city{Kharagpur}
 \country{India}}

\author{Rajdeep Mukherjee}
\affiliation{%
 \institution{Indian Institute of Technology}
 \city{Kharagpur}
 \country{India}}

\author{Niloy Ganguly}
\affiliation{%
 \institution{Indian Institute of Technology}
 \city{Kharagpur}
 \country{India}\\
 \institution{Leibniz University of Hannover}
 \city{Hannover}
 \country{Germany}
 }

% \author{Niloy Ganguly}
% \affiliation{%
%  \institution{Indian Institute of Technology Kharagpur, India}
%  \city{Leibniz University of Hannover}
%  \country{Germany}}

\author{Saptarshi Ghosh}
\affiliation{%
 \institution{Indian Institute of Technology}
 \city{Kharagpur}
 \country{India}}

%%
%% By default, the full list of authors will be used in the page
%% headers. Often, this list is too long, and will overlap
%% other information printed in the page headers. This command allows
%% the author to define a more concise list
%% of authors' names for this purpose.
% \renewcommand{\shortauthors}{Trovato and Tobin, et al.}

%%
%% The abstract is a short summary of the work to be presented in the
%% article.
\begin{abstract}
Convincing people to get vaccinated against COVID-19 is a key societal challenge in the present times. 
As a first step towards this goal, many prior works have relied on social media analysis to understand the specific concerns that people have towards these vaccines, such as potential side-effects, ineffectiveness, political factors, and so on. 
Though there are datasets that broadly classify social media posts into Anti-vax and Pro-Vax labels, there is no dataset (to our knowledge) that labels social media posts according to the specific anti-vaccine concerns mentioned in the posts. 
In this paper, we have  curated CAVES, the first large-scale dataset containing   
about 10k COVID-19 anti-vaccine tweets labelled into various specific anti-vaccine concerns in a multi-label setting.
This is also the first multi-label classification dataset that provides explanations for each of the labels.
Additionally, the dataset also provides class-wise summaries of all the tweets. 
We also perform preliminary experiments on the dataset and show that this is a very challenging dataset for multi-label explainable classification and tweet summarization, as is evident by the moderate scores achieved by some state-of-the-art models.
Our dataset and codes are available at: \url{https://github.com/sohampoddar26/caves-data}

\end{abstract}

%%
%% The code below is generated by the tool at http://dl.acm.org/ccs.cfm.
%% Please copy and paste the code instead of the example below.
%%
% \begin{CCSXML}
% <ccs2012>
%   <concept>
%       <concept_id>10003120.10003130</concept_id>
%       <concept_desc>Human-centered computing~Collaborative and social computing</concept_desc>
%       <concept_significance>500</concept_significance>
%       </concept>
%   <concept>
%       <concept_id>10002951.10003317.10003347.10003356</concept_id>
%       <concept_desc>Information systems~Clustering and classification</concept_desc>
%       <concept_significance>500</concept_significance>
%       </concept>
%   <concept>
%       <concept_id>10002951.10003317.10003347.10003357</concept_id>
%       <concept_desc>Information systems~Summarization</concept_desc>
%       <concept_significance>300</concept_significance>
%       </concept>
%  </ccs2012>
% \end{CCSXML}

% \ccsdesc[500]{Human-centered computing~Collaborative and social computing}
% \ccsdesc[500]{Information systems~Clustering and classification}
% \ccsdesc[300]{Information systems~Summarization}

% %%
% %% Keywords. The author(s) should pick words that accurately describe
% %% the work being presented. Separate the keywords with commas.
\keywords{COVID-19; anti-vaccine concerns; tweets; dataset; multi-label classification; explainable classification; summarization.}

%%
%% This command processes the author and affiliation and title
%% information and builds the first part of the formatted document.
\maketitle

\section{Introduction}
\label{sec:intro}

\blfootnote{*This work has been accepted to appear at SIGIR'22 (Resource Track)}
The COVID-19 pandemic has been ongoing since 2020 and has affected hundreds of millions of people till date. 
%There have been quite a few vaccines made by different pharmaceutical companies that have helped keep the spread of virus in check to some extent. 
Medical professionals believe that vaccination at a near-societal scale is the best way to achieve herd immunity and eradicate the virus~\footnote{\url{https://tinyurl.com/WHO-COVID-immunity}}.
% \todo{need citation}
However, unfortunately a significant fraction of population  are skeptical about taking COVID-19 vaccines due to different reasons~\cite{poddar2022winds, bonnevie2021quantifying}, and
%\st{even if they were not against vaccines prior to the COVID-19 pandemic}. 
understanding their concerns is very important for convincing such people about the benefits of the vaccines.

Concerns about vaccination is not new. Even in pre-Covid times, there have been many people who are known to be {\it anti-vax} or {\it vaccine-hesitant}.
Researchers have studied the concerns of such people towards vaccines, both through surveys as well as through the lens of social media like Twitter~\cite{paul2021attitudes,bonnevie2021quantifying}. 
% \todo{need citation}
Especially, understanding vaccine-related opinions through the lens of social media (e.g., Twitter) is a very popular method due to the large scale at which such studies can easily be conducted (as opposed to human surveys that are very difficult to conduct at large scale).
Several datasets have also been developed to aid the study of vaccine-related opinions on social media. For instance, there exist  datasets that label tweets (microblogs) into three broad categories of pro-vaccine (i.e., supporting vaccines), anti-vaccine (i.e., opposing vaccines) and neutral~\cite{muller2019crowdbreaks,cotfas2021longest}. 
However, when it comes to \textit{understanding the reasons behind vaccine hesitancy} (i.e., a finer analysis of anti-vaccine concerns), researchers have only relied on human surveys / manual analysis at a small scale or unsupervised topic models which do not perform too well and require manual intervention. 
To the best of our knowledge, there does {\it not} exist any large-scale dataset of social media posts \textit{labeled with various reasons of vaccine-hesitancy}.

In this paper, we provide the first such large-scale dataset by annotating tweets into these fine-grained concerns about vaccines. 
We first extensively collect tweets (via the Twitter API) using more than 200 keywords related to vaccines, and then employ a 3-class classifier to identify tweets that are very likely to be anti-vaccine.
Subsequently we get such tweets labeled (by three human annotators) according to the specific anti-vaccine concern(s) stated in the tweet-text. 
The final dataset -- which we call \textbf{C}oncerns \textbf{A}bout \textbf{V}accines with \textbf{E}xplanations and \textbf{S}ummaries (CAVES) -- contains about 10K anti-vaccine tweets related to COVID-19 vaccination, which are manually labeled with 11 established anti-vaccine concerns, such as concerns about the vaccine ingredients,  side-effects of vaccines, and so on (see Table~\ref{tab:classes} for a list of these concerns). 
Since a particular tweet often reflects multiple anti-vaccine concerns, the dataset is \textit{multi-labeled}.
Additionally, the dataset is curated with  
human-annotated \textit{explanations} for the labels, indicating which exact parts of the tweet-text indicates a particular anti-vaccine concern. 

Once these classes are identified, it also makes sense to summarize the tweets in a class, so that the authorities can quickly glance through the various concerns being raised by people and act accordingly to promote the use of vaccines. To this end, the CAVES dataset contains
human-compiled summaries of the tweets pertaining to each anti-vaccine concern (i.e., class-wise summaries).

The CAVES dataset is very timely in the present pandemic situation, and can help social media researchers / authorities gain a fine-grained understanding of anti-vaccine concerns among masses. %\noteng{removed sentences from here}
%Additionally, the dataset is novel from the perspective of machine learning in the following way. Our dataset facilitates multi-label classification while providing explanations of each label from the text. While multi-label classification or  explanations for labels is not new, there hardly exists any dataset (to our knowledge) that provides explanations in a multi-label setting, with \textit{separate explanations for each label} associated with a text.We also provide the class-wise abstractive summaries which can be used to evaluate summarization models~\cite{liu2012graph,liu2019hierarchical}. The class labels and summaries together can also be used to develop and evaluate classification-cum-summarization models~\cite{rudra2015extracting,roy2020classification}. 
%
%
%These models can prove beneficial to quickly understand a summary of a particular concern of multiple people over some demographic (say during a particular time-period or a particular geographical location). Thus our dataset has three primary use cases, of multi-label classification, generating explantions for each label and summarization of classes
There are three major use-cases of the CAVES dataset  -- 
(i)~the dataset can be used to develop models for automated fine-grained retrieval and analysis of concerns people have towards vaccines;
(ii)~the dataset can act as an excellent resource for performing the dual tasks of  multi-label classification and explanation identification (where separate explanations for each of the different labels are provided),
and 
(iii)~the summaries provided for each class would act as an excellent resource %dataset also contains the class-wise summaries 
for performing the task of multi-document or tweet-stream summarization.
% For the purpose of review, a data sample has been provided; see Section~\ref{sub:availability} for details. 
%The complete dataset will be released upon acceptance of the paper.

We benchmark the CAVES dataset with respect to the three tasks stated above (multi-label classification, explanation generation, summarization). For each task, we apply several state-of-the-art methods.
We demonstrate that the CAVES dataset is quite challenging for these tasks. 
For instance, in the multi-label classification task, the best performance achieved is a moderate Macro-F1 of $0.6$, while the performances in the explanation generation task is even lower.
Hence, there is a need for better models for these tasks.
We have provided the dataset along with these benchmarking codes on Github (see Section~\ref{sub:availability} for details).

\section{Literature Review}
\label{sec:litsurvey}

In this section we shall briefly touch on some previous works that have tried to understand the concerns about vaccines, highlighting the importance of our dataset. Additionally we briefly discuss some of the other datasets available for the tasks of multi-label classification, explanation generation and summarization.

\subsection{Concerns about Vaccines}
\label{sec:relwork_concerns}

Several prior works have been trying to understand people's stance towards vaccines from social media like Twitter. 
One of the largest datasets about vaccine stances is that by \citet{muller2019crowdbreaks} which contains about 28K tweets labelled into three different categories -- i)~Anti-Vax (tweets that are against vaccines), ii)~Pro-Vax (tweets that support vaccines) and iii)~Neutral (other tweets that talk about vaccine without a clear stance).
\citet{yuan2019examining} provides another dataset that labels tweets in a similar fashion.
After the onset of the COVID-19 pandemic the discourse around vaccines have risen a lot, especially hesitancy towards vaccines~\cite{johnson2020online,bonnevie2021quantifying,poddar2022winds}. Naturally, researchers have tried to build labelled datasets for vaccine-stance detection using COVID-19 vaccine related tweets~\cite{cotfas2021longest,poddar2022winds}.
However, these datasets only provide the broad-level classification of tweets (pro-vax, anti-vax and neutral) and do not deal with the particular reasons for vaccine hesitancy.

Several researchers have tried to understand the concerns of people causing the rise of vaccine hesitancy during COVID, usually with the help of surveys~\cite{troiano2021vaccine,paul2021attitudes,sonawane2021covid,dhama2021covid,nuzhath2020covid}.
Some works have also examined Twitter posts to understand these concerns, through manual analysis of a small-sample~\cite{bonnevie2021quantifying}, or by applying topic models such as LDA~\cite{praveen2021analyzing} or through a combination of both~\cite{poddar2022winds}. 
However, there exists no large-scale labelled dataset that Machine Learning models can leverage to perform automated detection of these fine-grained concerns towards vaccines.

\subsection{Multi-Label Classification, Explanation Generation and Summarization}

Multi-Label text classification is a classical machine learning problem with several datasets available for it in different domains. Reuters dataset~\cite{lewis2004rcv1} is one of the oldest multi-label datasets containing news articles which are labelled into different categories. %based on regions, topics and industries.
The MIMIC-III dataset~\cite{johnson2016mimic} is a large dataset of medical records labelled into several ICD-9 codes. 
There also exist a few datasets performing multi-label classification on tweets, e.g. SemEval 2018 Task 1 dataset~\cite{mohammad2018semeval} containing several sub-classes of emotions, and TREC-IS dataset~\cite{mccreadie2019trec} which contains different label categories of disaster related tweets.

In the past few years, explainable machine learning models have been an important research area and there have been several models trying to provide explanations~\cite{ribeiro2016should}. 
%One very popular explanation generation method is LIME~\cite{ribeiro2016should}.
There also exist quite a few datasets  that provide explanations/rationales for classification. A popular collection of datasets is the ERASER benchmark~\cite{deyoung2020eraser} which comprise of several datasets for providing explanations for different tasks. Another dataset from the domain of Hate-speech detection is HateXplain~\cite{mathew2021hatexplain} which provide explanations for different tweets being hateful/offensive.
Though there exist some methods that can generate explanations for multi-label classification~\cite{li2021heterogeneous,mullenbach2018explainable}, to the best of our knowledge, {\it there exists no dataset that provides explanations in a multi-label setting, with separate explanations for different labels}. 
The CAVES dataset developed in this paper is the first dataset containing such distinct explanations for each label assigned to a text (tweet).

%One of the most popular summarization datasets is the CNN/Daily Mail dataset~\cite{nallapati2016abstractive} containing news articles along with their headlines (summaries). Another very popular dataset is the DUC-2004 data~\footnote{\url{https://www-nlpir.nist.gov/projects/duc/intro.html}}.
Multi-document summarization, as the name suggests deals with summarization of information contained in multiple documents. One such popular dataset is the Multi-News dataset~\cite{fabbri2019multi}. Summarization of tweets also falls under this category, and there exist a few different datasets containing summaries of tweets from different categories/domains of tweets.
For example, TGSum~\cite{cao2016tgsum} contains summaries of news-related tweets, while TSix~\cite{nguyen2018tsix} deals with summaries tweets about some real-world events. \citet{he2020tweetsum} provides another dataset of summaries of events from Twitter, while \citet{rudra2015extracting} and \citet{dutta-ensemble-summarization} provide datasets containing summaries of disaster-related tweets.
However, to our knowledge, there is no existing dataset focusing on summarization of general public opinions about COVID-19 vaccines. 
\section{Dataset Preparation}
\label{sec:dataset}

% \todo{why have we taken some decisions?}

In this section, we describe the CAVES dataset and its preparation in detail. 
%how we fetched the data from Twitter, selected the classes and got them annotated. 

%%%%%%%%%%%%%%%%%%%%%%%%%%%%%%%%%%%%%%%%%%%%%%%%%%%%%%%%%%%%%%%%%%%%%%%%%%
\subsection{Selection of tweets}
\label{sec:tweet_selection}

\vspace{1mm}
\noindent \textbf{Fetching tweets:}
We fetched tweets using the official Twitter API with various keywords related to vaccines in general (e.g., `vaccine', `vaxxer')  and COVID-19 vaccines in particular (e.g., names of COVID vaccines and their manufacturers such as `comirnaty', `covishield', `moderna', `gamaleya'). 
Some sample keywords are stated in Table~\ref{tab:data_keywords}.
We also added the set of 126 Anti-Vaccine and 154 Pro-Vaccine hashtags provided by~\citet{gunaratne2019temporal} (in the supplemental material of their work).
Using all of these keywords, we collected about 100M distinct vaccine-related tweets (excluding retweets) 
posted in between January 2020 and October 2021.

\begin{table}[!t]
    \centering
    %\footnotesize
    % \small
    \caption{Some of the keywords used for collection of tweets.}
    \label{tab:data_keywords}
    \begin{tabular}{|p{80mm}|}
    \hline
    \textbf{Generic keywords:} 
    vaccine, vaxxer, vaxxed, vaccinated, vaccination, covid vaccine, corona vaccine \\
    \hline 
    \textbf{COVID vaccine-specific keywords:} 
    astra zeneca, novavax, pfizer, BioNTech, comirnaty, moderna, gamaleya, NIAID, bharat biotech, covaxin, covishield, sanofi, curevac, sinovac, sinopharm, janssen, oxford vaccine, johnson vaccine, russian vaccine, chinese vaccine\\
    \hline
    \end{tabular}
    % \vspace{-8mm}
\end{table}

% Using this method of filtering tweets we were left with \todo{XXXX} tweets, which are very likely to be vaccine hesitant.
% Even after removing retweets there were quite a lot of similar tweets remaining, e.g. tweets that contain the same text but a few different mentions or hashtags.
% Such near duplicate tweets were also removed by keeping only one copy of tweets that had more than 80\% token overlap after removing the mentions and hashtags, as described in previous works~\cite{tao2013groundhog} 
% \noteng{was this a good idea?}

\vspace{2mm}
\noindent \textbf{Extracting Anti-Vax tweets:}
In this work, we specifically wanted to deal with anti-vaccine / anti-vax tweets that exhibit hesitancy towards vaccines. 
Such tweets make up only a small fraction of all vaccine-related tweets, and selecting a complete random sample would lead to only a small amount of anti-vax tweets. 
Hence, we decided to specifically identify anti-vax tweets using a COVID-vaccine stance classifier developed in our prior work~\cite{poddar2022winds}. 
This BERT-based classifier gives the probability of a given tweet being Anti-Vax, Pro-Vax and Neutral. 
The classifier was tested over multiple datasets containing vaccine-related tweets (details in~\cite{poddar2022winds}) and achieved macro-F1 scores in the range of [0.78, 0.825].
This classifier was first run over all the tweets that we had collected to obtain the probability of the tweets being Anti-Vax.
To be certain that we are left with a set of mostly anti-vax tweets, after testing several thresholds, we selected the tweets which the classifier predicted Anti-Vax with high confidence, i.e. with probability $\geq 0.8$ (similar thresholds have been used in prior works~\cite{mitra2016understanding} to filter out Anti-vax tweets with high precision).
Finally we were left with about 7.5M tweets which are highly likely to be vaccine-hesitant (according to the classifier). 
%We will be providing the tweet IDs of all these tweets with our dataset.
Through the annotation study described later in Section~\ref{sec:annotation_process}, we observe that 98\% of the tweets that are predicted to be anti-vaccine with probability $\geq 0.8$ are actually anti-vaccine.

% Finally We use a random sample from this set of vaccine-hesitant tweets for further analysis.

\begin{table}[tb]
    \centering
    % \footnotesize
    \small
    \caption{Mapping the different anti-vaccine concerns given by previous works to our set of concerns (classes).}
    \label{tab:class_coverage}
    \begin{tabular}{|l||p{20mm}|p{20mm}|p{17.5mm}|}
    \hline
    \textbf{Our classes} & \textbf{Praveen~\cite{praveen2021analyzing}} & \textbf{Nuzhath~\cite{nuzhath2020covid}} & \textbf{Bonnevie~\cite{bonnevie2021quantifying}}\\
    \hline \hline
    Conspiracy & - & Unusual theories & - \\
    \hline
    Country & Skepticism over the nationality of the vaccine  & - & - \\
    \hline
    Ineffective & - & Vaccine will be Ineffective & - \\
    \hline
    Ingredients & - & - & Vaccine Ingredients \\
    \hline
    Mandatory & - & Freedom of choice & - \\
    \hline
    Pharma & Negative feeling towards pharma companies  & Profit from developing a COVID-19 vaccine; Mistrust of Scientists and vaccine advocates & Pharmaceutical Industry; Federal Health Authorities \\
    \hline
    Political & - & Mistrust in the government & Policies \& Politics \\
    \hline
    Religious & - & Religious beliefs & Religion \\
    \hline
    Rushed & Skepticism over vaccine trails; Doubts regarding data; Rush in providing the vaccine & Vaccines are untested; Fast paced Vaccine Development & Research \& Clinical Trials \\
    \hline
    Side-effect & Fear over health; Allergic reactions; Fear of death & Vaccine will have side effects; Vaccines cause illnesses of unknown origin & Negative health impacts; Vaccine Safety \\
    \hline
    Unnecessary & COVID-19 being exaggerated & Vaccine is Unnecessary & Disease Prevalence \\
    \hline
    \end{tabular}
    % \vspace{-6mm}
\end{table}

\begin{table*}[!ht]
    \centering
    % \footnotesize
    \small
    \caption{Description of different concerns towards vaccines. along with an example tweet (excerpts). These concerns define the classes in our dataset. The keywords marked in bold in the tweets are the explanations identified by annotators. The percentage of tweets in each class (last column) is calculated with respect to the total number of tweets in the final dataset.}
    \label{tab:classes}
    \begin{tabular}{|p{9.5cm}|p{5.5cm}|c|}
    \hline
        \textbf{Class Description} & \textbf{Example Tweet Excerpt} & \textbf{\#Tweets} \\
    \hline \hline
        \textbf{Unnecessary} - The tweet indicates COVID is not dangerous, vaccines are unnecessary, or that alternate cures (such as hydroxychloroquine) are better. & 
        I wouldn't get a vaccine for \textbf{a virus w/a 95\% recovery rate} &
        722 (7.3\%)\\
        \hline

        \textbf{Mandatory} - The tweet is against mandatory vaccination and talks about their freedom. &
        No one \textbf{should be forced} to get a vaccine! \textbf{\#medicalfreedom}  &
        783 (7.9\%)\\
        \hline

        \textbf{Pharma} - The tweet indicates that the Big Pharmaceutical companies are just trying to earn money, or is against such companies in general because of their history. &
        % Hydroxychloroquine is proven but \textbf{Moderna and Pfizer won't make money} with it. 
        \textbf{Pfizer} is about to unleash hell in the name of \textbf{profit} \& reckless endangerment.
        &
        1273 (12.8\%) \\
        \hline

        \textbf{Conspiracy} - The tweet suggests some deeper conspiracy, e.g., vaccines are being used to track people via microchips, the entire COVID is a hoax, vaccines are used for behaviour and mind control, vaccines alter DNA, vaccines are bio-weapons, etc. (list of conspiracy theories compiled from~\cite{nuzhath2020covid,shahsavari2020conspiracy}) &
        % \todo{CHANGE} Bill Gates has to design a \textbf{smaller microchip} &
        % @thatsmanderley @nicolatsproston @MattHancock They'll be a vaccine, 
        it won't be a covid vaccine though! 
        Remember \textbf{depopulation, smart dust microchips}? It's what Bill Gates had planned all along!! &
        487 (4.9\%) \\
        \hline

        \textbf{Political} - The tweet expresses concerns that the governments / politicians are pushing their own agenda though the vaccines. &
        It took \textbf{Donald Trump} to turn me into an anti-vaxxer. &
        626 (6.3\%)\\
        \hline

        \textbf{Country} - The tweet is against some vaccine because of the country where it was developed / manufactured  &
        I do not want a \textbf{EU crappy} mRNA vaccine! &
        201 (2.0\%) \\
        \hline

        \textbf{Rushed} - The tweet expresses concerns that the vaccines have not been tested properly, have been rushed or that the published data is not accurate. &
        \textbf{no phase 3 trials}, Brazil rejected Covaxin &
        1,477 (14.9\%) \\
        \hline

        \textbf{Ingredients} - The tweet expresses concerns about the ingredients present in the vaccines (eg. fetal cells, chemicals) or the technology used (e.g., mRNA) &
        \textbf{nanoparticles} in Pfizer’s vaccine trigger rare allergic reactions &
        436 (4.4\%) \\
        \hline

        \textbf{Side-effect} - The tweet expresses concerns about the side effects of the vaccines, including deaths caused. &
        did u hear about the Johnson and Johnson vaccine \textbf{blood clots}? &
        3,805 (38.4\%) \\
        \hline

        \textbf{Ineffective} - The tweet expresses concerns that the vaccines are not effective enough and are useless. &
        They rushed into AZ, despite its \textbf{effectivity being 20\% less} &
        1,672 (16.9\%) \\
        \hline

        \textbf{Religious} - The tweet is against vaccines because of religious reasons &
        \textbf{Bishops discourage Catholics} from receiving Johnson \& Johnson vaccine &
        64 (0.6\%) \\
        \hline
 
        \textbf{None} - The tweet states no clear reason for vaccine hesitancy, or any of the other reasons. &
        They should be offered a free bullet with the vaccine &
        629 (6.3\%) \\
        % \hline \hline
    % \textbf{Neutral} - The tweet does NOT indicate hesitancy towards any vaccine. &
        % I had pfizers, arm was sore. But who cares, it's just arm soreness. \\
    \hline
    \end{tabular}
    % \vspace{-8mm}
\end{table*}

\vspace{2mm}
\noindent {\bf Removing duplicates for annotation:}
Even after removing the retweets, 
% \noteng{line above going outside the margin} 
there were quite a lot of similar tweets remaining, e.g. tweets that contain the same text but a few different mentions or hashtags.
For the human annotation (described in Section~\ref{sec:annotation_process}), we removed such near-duplicate tweets
using some of the methods described in~\cite{tao2013groundhog}.
We measured the similarity between two tweets by their token overlap after removing the mentions and hashtags.
We retain only one copy of tweets that had more than 80\% token overlap~\cite{tao2013groundhog}.
After this step, we are left with around 6.5M anti-vax tweets.

%%%%%%%%%%%%%%%%%%%%%%%%%%%%%%%%%%%%%%%%%%%%%%%%%%%%%%%%%%%%%%%%%%%%%%%%%%
\subsection{Selection of anti-vaccine concern classes}
\label{sec:data_classes}

\vspace{1mm}
\noindent \textbf{Defining the classes:}
An important decision was the selection of the classes in our dataset, i.e., the specific anti-vaccine concerns.
We initially consulted various prior works that curate lists of anti-vaccine concerns~\cite{paul2021attitudes,bonnevie2021quantifying,praveen2021analyzing,nuzhath2020covid,poddar2022winds,dhama2021covid} (see Section~\ref{sec:relwork_concerns}). %These can include different classes such as ``Concerns about commercial profiteering''~(\cite{paul2021attitudes}), ``Negative health impacts''~(\cite{bonnevie2021quantifying}), etc.
The classes provided by a few such prior works are listed in the second, third and fourth columns of Table~\ref{tab:class_coverage}.
We observed that different prior works have considered different sets/classes of anti-vaccine concerns. 
The classes also vary in granularity -- multiple classes considered in one study can be combined into a single class considered in another study.  %For example, three classes described in~\citet{praveen2021analyzing} -- `\textit{fear over health}', `\textit{allergic reactions}' and `\textit{fear of death}' -- can be combined into a single class `\textit{Health side-effects}'. 
Hence we decided to curate our own set of classes to appropriately capture the different concerns expressed by the tweets towards vaccines.

To this end, two of the authors examined the concern-classes listed in several prior works that attempted to categorize anti-vaccine concerns. 
They also manually examined a random sample of 500 tweets in 3 iterations to understand the different types of concerns voiced in the tweets. 
Based on extensive discussions after each iteration, a list of \textit{12 classes of anti-vaccine concerns} was finalized, that encompass the classes defined by most prior works as well.
The 12 classes are described in Table~\ref{tab:classes}.
Note that the `None' class is meant for tweets that oppose vaccines but do {\it not} give any reason / concern for such opposition.
%These 12 classes cover all the concerns that were put forward by different prior works. 
Table~\ref{tab:class_coverage} gives a mapping of how these 12 classes correspond to the classes defined by a few of the prior works.

\vspace{2mm}
\noindent \textbf{Multi-label setting with explanations:}
While manually examining the tweets, we noticed that several tweets actually express multiple concerns about vaccines; for instance, a tweet may say that the vaccines are insufficiently tested (``\textbf{Rushed}'') and thus can cause unknown complications (``\textbf{Side-effects}''). 
Some example tweets which talk about multiple concerns are given in Table~\ref{tab:multilabelexample}. 
Hence, we decided to label the tweets in a multi-label setting, assigning one or more classes to each tweet. This multi-label nature of the dataset can also be seen from the joint label distribution in Figure~\ref{fig:label_joint_dist} which is explained later in Section~\ref{sec:finalizing_labels}.

Another consequence of the multi-label nature of the tweets is that different parts of the tweet-text explain the different labels/classes assigned to a tweet (see the examples given in Table~\ref{tab:multilabelexample}). 
Since we want the dataset to support explainable classification, 
it is imperative for classifiers to get additional information to correctly identify the parts of the tweet that explain every label that is assigned to the tweet. 
Hence we also decided to get the explanations marked for each of the labels separately (and not just one explanation for the entire tweet).

% \todo{dont mention "Neutral", just say annotators removed.. even though classifier, some tweets could still..}
% We also decided to include a ``\textit{Neutral}'' class for the purpose of tweet annotation to include tweets which were falsely selected by the classifier as vaccine-hesitant. However, tweets marked with this class were later removed from the final dataset.

% \todo{conspiracy class- just we have taken some populat conspiracy theories as in table..}
% The ``\textit{conspiracy}'' class is a unique class since detection of different conspiracy theories is a different task altogether. During the initial analyses, the authors had some disagreement in what was considered as a conspiracy and what was not. Thus we decided to prepare a list of popular conspiracy theories from prior works~\todo{cite} as given in Table~\ref{tab:conspiracies}. We decided to label a tweet as ``\textit{conspiracy}'' if it had references to one of these conspiracies. Note that in most of such cases, the tweet also could be mapped to another class. For example, the ``\textit{tracking chips}'' conspiracy can also be labelled into the ``\textit{ingredients}'' class, and ``\textit{pandemic is fake}'' conspiracy can be labelled as the vaccine being ``\textit{unnecessary}''.

%manual analysis, previous works, description of classes, examples
% \todo{other reasons??}
% \todo{multi-label nature of the  dataset. talk about many tweets being multilabel}

%%%%%%%%%%%%%%%%%%%%%%%%%%%%%%%%%%%%%%%%%%%%%%%%%%%%%%%%%%%%%%%%%%%%%%%%%%
\subsection{Annotation of the tweets}
\label{sec:annotation_process}
% \todo{DISCUSS: cogitotech demographics, agreement scores??}

%\noteng{DONE. I think this section needs a lot more elaboration. Who are the annotators, how many annotators, were they given some initial coaching, were they provided some interface whereby they annotated. Did you organize the tweets as per class using an initial classifier. How long did the company take. How much was paid. -- every finer details needs to be given. what exact direction was given regarding the explanation. Is there an instruction manual - we need to publish that, were they given an idea that multi-label is generally sparse}
%\noteng{DONE: why 11K? How have you chosen them. }

\noindent \textbf{Annotation setup:}
We took the help of a data annotation firm named Cogitotech (\url{https://www.cogitotech.com/}) to get the tweets annotated. 
We chose to employ this company instead of a crowdsourcing platform (such as Amazon Mechanical Turk), since we wanted to have discussions with the annotators to properly describe the different classes.

We first provided an instruction manual to the firm containing the description of the different class labels (anti-vaccine concerns) along with a couple of examples (as given in Table~\ref{tab:classes}). 
Specifically, we provided the annotators the following instructions.
\textbf{(I1):}~For each tweet, select the labels based on the concern(s) that the tweet indicates towards COVID-19 vaccines. You can select more than one label if the tweet states multiple concerns towards vaccines.
\textbf{(I2):}~For each of the selected labels for a particular tweet, 
mark the specific part of the tweet that made you select that label. That is, mark a few keywords or small phrases from the tweet that explain your selection.
\textbf{(I3):}~If a tweet contains a URL, you can visit the content of the URL for better understanding of the tweet.
\textbf{(I4):}~Although we have tried to automatically select only tweets that are hesitant towards COVID vaccines, there may be some tweets that do not show any hesitancy towards the vaccines. Indicate such tweets separately by selecting the option ``The tweet is NOT hesitant towards any vaccines'', and do not assign any label to such tweets. Keywords need not be provided for such tweets. 
\textbf{(I5):}~If a tweet opposes vaccines but does {\it not} give any reason for such opposition, or gives some reason other than the concerns we have listed, then select the `None' label. Keywords need not be provided for such tweets.

%\noindent $\bullet$ \textbf{I6:} Select the ``\textbf{Conspiracy}'' label if it contains some claim which is similar to the ones listed in Table~\ref{tab:conspiracies}.  

%\vspace{1mm}
%We specifically mentioned about the conspiracy class since we found that there were some scope for confusion as
%the conspiracies can often be labelled into other classes too -- e.g. mention of ``tracking chips'' can be also labelled into the ``Ingredients'' class.
%The list of conspiracy theories given in Table~\ref{tab:conspiracies} was compiled from some previous works~\cite{nuzhath2020covid,shahsavari2020conspiracy} and the MediaEval 2021 Fake News shared task~\footnote{\url{https://github.com/konstapo/2021-Fake-News-MediaEval-Task/}}.
%\noteng{DONE. I think this instruction is still not looking good, is there a conspiracy guide which we can provide, because controversial is a vague word}

\vspace{1mm}
\noindent \textbf{Annotation Process:}
We asked the firm to get each tweet annotated independently by four annotators. 
The annotators were a set of university graduates of the age group 20-30, who are well-versed in English and are conversant with Twitter.
Each tweet was labeled by four annotators from this set.

We first provided a set of 1,000 randomly sampled tweets (from the set of tweets obtained after duplicate removal, as described in Section~\ref{sec:tweet_selection}) to the firm to be annotated independently by four annotators. 
After a week, the firm completed the annotations and we cross-checked the annotations for 100 of these tweets for correctness.
Other than a few minor corrections to the labels, the annotations looked good. 
We had a discussion with the annotators and clarified some of their doubts regarding the class descriptions. We also clarified that explanations (for a particular label) within a particular tweet can consist of non-contiguous words as well.

Once we were satisfied with the annotations, we then provided them another set of randomly sampled 10K tweets to be annotated.
The firm agreed to annotate the total of 11k tweets for about INR~110,000 ($\sim$USD~1500), and completed the annotations in about 2 months. 
The tweets were marked by different annotators in their firm, with four annotators marking each tweet.

%%%%%%%%%%%%%%%%%%%%%%%%%%%%%%%%%%%%%%%%%%%%%%%%%%%%%%%%%%%%%%%%%%%%%%%%%%
\subsection{Finalizing the labels}
\label{sec:finalizing_labels}

\begin{table}[tb]
    \centering
    \small
    \caption{Examples of Multi-Label tweets. The explanations for the three labels are highlighted in \textcolor{blue}{\emph{blue}}, \textcolor{red}{\it red} and \textcolor{brown}{\bf brown}. The overlaps between explanations are highlighted in \textcolor{cyan}{\underline{cyan}}.}
    \label{tab:multilabelexample}
    \begin{tabular}{|p{6cm}|p{1.5cm}|}
    \hline
    \textbf{Tweet excerpt} & \textbf{Labels} \\
    \hline
    \textcolor{blue}{\emph{No claims, no trials}}, no compensation: \textcolor{red}{\it Pfizer given protection} from legal action by \textcolor{brown}{\bf UK government} & \textcolor{blue}{\emph{Untested}}, \textcolor{red}{\it Pharma}, \textcolor{brown}{\bf Political} \\
    \hline
    Good only to \textcolor{cyan}{\underline{implant their chips}}, of course that is the entire purpose... study says \textcolor{brown}{\bf jab ineffective} against local variant & \textcolor{blue}{\emph{Conspiracy}}, \textcolor{red}{\it Ingredients}, \textcolor{brown}{\bf Ineffective} \\
    \hline
    \textcolor{blue}{\emph{Johnson \& Johnson}}, another company at the top of the broken \textcolor{blue}{\emph{Big Pharma Ecosystem}}, has created a vaccine that was \textcolor{red}{\it rapidly approved}. & \textcolor{blue}{\emph{Pharma}}, \textcolor{red}{\it Untested}\\
    \hline
    It's just another annual new corona virus. It's \textcolor{blue}{\emph{on its way out}}. Talk of a vaccine being pushed by a certain \textcolor{red}{\it billionaire computer nerd} & \textcolor{blue}{\emph{Unnecessary}}, \textcolor{red}{\it Pharma}\\
    \hline
    \end{tabular}
    % \vspace{-8mm}
\end{table}

\begin{figure}[tb]
    \centering
    \includegraphics[width=\linewidth]{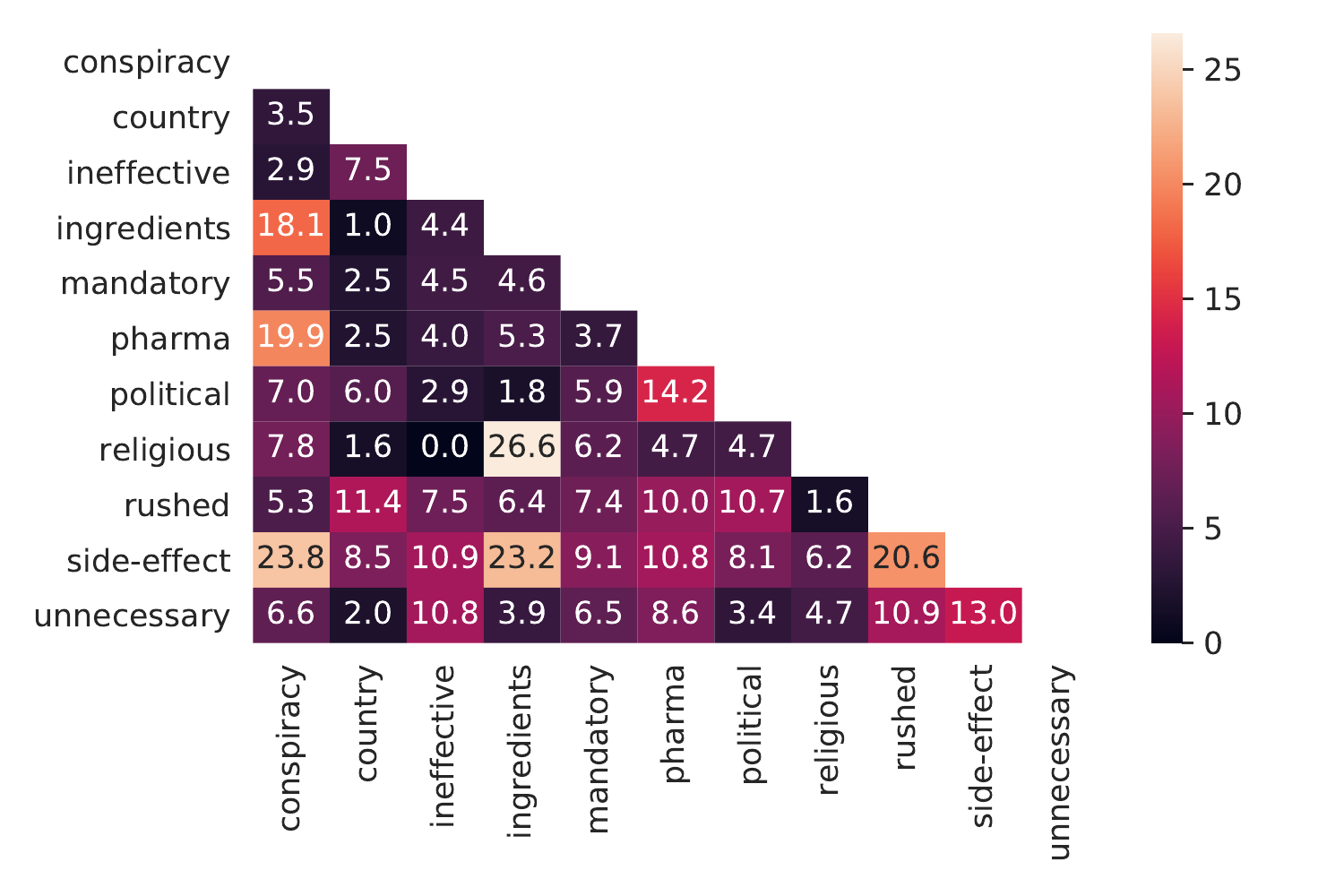}
    % \vspace{-4mm}
    \caption{Joint distribution of different classes. For a pair of classes $c_i$ and $c_j$, the value in a cell is $n_{ij} \times 100 / \min \{|c_i|, |c_j|\}$, where $n_{ij}$ is the number of tweets that have been labeled with both $c_i$ and $c_j$.}
    \label{fig:label_joint_dist}
    \Description{This figure shows the joint distribution of different classes in the overall dataset. Each number in a cell represents the percentage of overlap divided by the maximum possible overlap between the two classes. Some pairs of classes are more correlated with each other.}
    % \vspace{-5mm}
\end{figure}

Out of the 11k tweets we got annotated, 226 tweets were deemed to \textit{not} express any hesitancy towards vaccines (i.e., these tweets were misclassified by the pro-vaccine / neutral / anti-vaccine classifier  described in Section~\ref{sec:tweet_selection}), and were thus removed from the dataset. 

Recall that four annotators independently assigned one or more labels to each tweet.
Taking a union of all the labels for each tweet, we were left with about 5,600 tweets with a single label, 4,200 tweets with two distinct labels, 900 tweets with three distinct labels, and 100 tweets with more than three distinct labels. 
However, some of these tweets were marked with a particular label by only one annotator out of the four. 
On manually examining some of these tweets and labels, we found a few cases where the label did \textit{not} accurately represent the concerns in the tweets, and probably represented personal opinions.
%Thus the unions of the labels cannot be considered with certainty.
Thus, for each tweet, \textit{we consider only those labels that are given by at least two annotators}, since such labels are more likely to represent the concerns expressed in the tweet. 
For example, suppose a tweet has been labelled by the four annotators ($A1-A4$) as follows -- $A1:\{C1, C2, C3\}$; $A2:\{C2, C3, C4\}$;  $A3:\{C3\}$;  $A4:\{C2, C5\}$ (where $C1-C5$ represent some classes), then the final ground truth labels for the tweet will be $L:\{C2, C3\}$, since these labels have been selected by at least two annotators. We then removed 853 tweets that did not have any label assigned by at least two annotators.

Using this strategy we were left with 9,921 tweets, out of which 1716 have exactly two distinct labels, 269 have three distinct labels, and the rest have a single label. No tweet has more than 3 labels.
% A few examples of tweets with multiple labels is given in Table~\ref{tab:multilabelexample}. 
The distribution of labels class-wise is given in Table~\ref{tab:classes} (last column).

% \new{Note that we did not calculate the inter-annotator aggrement (IAA) since all the tweets were each labelled by 3 people (not just a subset of the tweets), and since the 3 different labels given per tweet were not marked by 3 distinct people, but rather a group of people (similar to a crowdsourced setting).
% }

To get an idea of which classes frequently occur with some other classes,
we have also calculated the joint-distribution of classes (shown in Figure~\ref{fig:label_joint_dist}). 
Let $c_i$ and $c_j$ be two classes. We check the number $n_{ij}$ of tweets that have been labeled with both $c_i$ and $c_j$, and divide this number by the maximum number of possible tweets where both classes could have been present (i.e. the number of tweets  in the smaller class out of $c_i$ and $c_j$).
This distribution of $n_{ij} \times 100 / \min \{|c_i|, |c_j|\}$ is given in Figure~\ref{fig:label_joint_dist}. It can be seen that some particular classes co-exist frequently with other classes. For example, the `Conspiracy' class often co-occurs with the `Ingredients', `Pharma', and `Side-effect' classes. Similarly, the `Ingredients' and `Rushed' classes also co-exist often with the `Side-effect' class.

%%%%%%%%%%%%%%%%%%%%%%%%%%%%%%%%%%%%%%%%%%%%%%%%%%%%%%%%%%%%%%%%%%%%%%%%%%
\subsection{Finalizing the explanations}

\begin{table}[tb]
    \centering
    % \footnotesize
    \small
    \caption{Sample explanations given by different annotators. The tweets marked by the three annotators are highlighted in \textcolor{blue}{\emph{blue}}, \textcolor{red}{\it red} and \textcolor{brown}{\bf brown}. The overlaps are highlighted in \textcolor{cyan}{\underline{cyan}}.}
    \label{tab:explanations_example}
    \begin{tabular}{|l|p{6.5cm}|}
    \hline
    \textbf{Class} & \textbf{Tweet} \\ 
    \hline \hline
    Side-effect &
    \textcolor{red}{\it 53 dead in Gibraltar in 10 days} after experimental Pfizer mRNA COVID injections started. Could the new vaccines be \textcolor{blue}{\emph{causing all those ``COVID deaths''}}? Tiny Gibraltar Shines Huge Light on \textcolor{brown}{\bf Vaccine Deaths}.\\
    \hline
    Ineffective &
    Get vaxxed with \textcolor{red}{\it Pfizer} \textcolor{cyan}{\underline{with 39\% efficacy}} or with J\&J with even much lower efficacy so you can spread virus without testing. \\
    \hline
    Pharma & 
    Somebody is \textcolor{red}{\it going to} \textcolor{cyan}{\underline{make money}} \textcolor{blue}{\emph{from all of this}}, why not let it be you? Find the \textcolor{brown}{\bf Big Pharma companies that are front runners} for creating the vaccine, which would = huge \$\$\$. \\
    \hline
    Ineffective &
    @USER \textcolor{red}{\it You can't actually say it} \textcolor{cyan}{\underline{will save}} \textcolor{blue}{\emph{others or even those vaccinated.}} \\
    \hline
    \end{tabular}
    % \vspace{-6mm}
\end{table}

\begin{figure}[!t]
    \centering
    \begin{subfigure}[b]{0.47\linewidth}
         \centering
         \includegraphics[width=\linewidth, height=4cm]{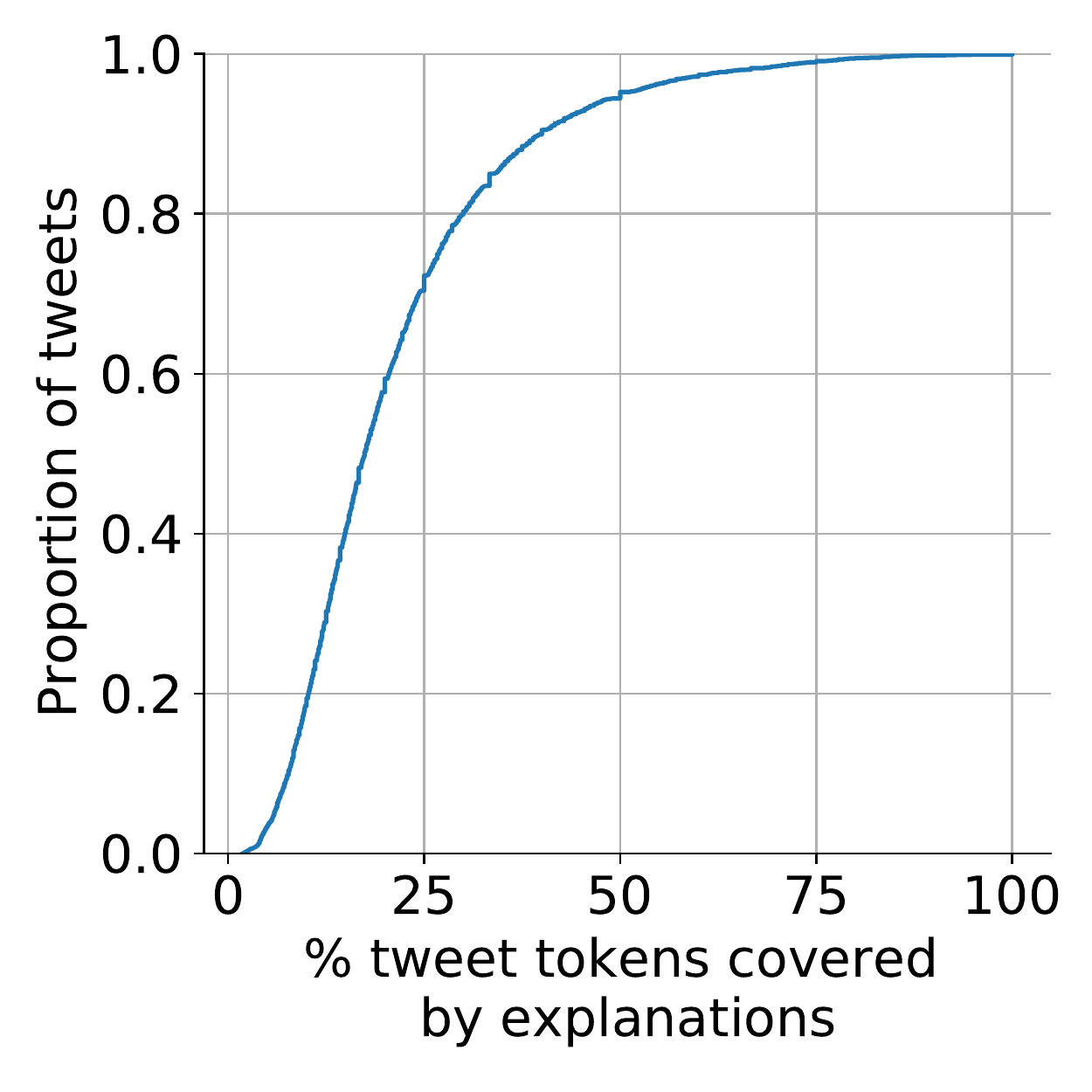}
         \caption{CDF of \% of tweet tokens covered by explanations, after taking union of keywords selected by individual annotators}
         \label{fig:keyworda}
     \end{subfigure}
     \;\;\;
    \begin{subfigure}[b]{0.47\linewidth}
         \centering
         \includegraphics[width=\linewidth,height=4cm]{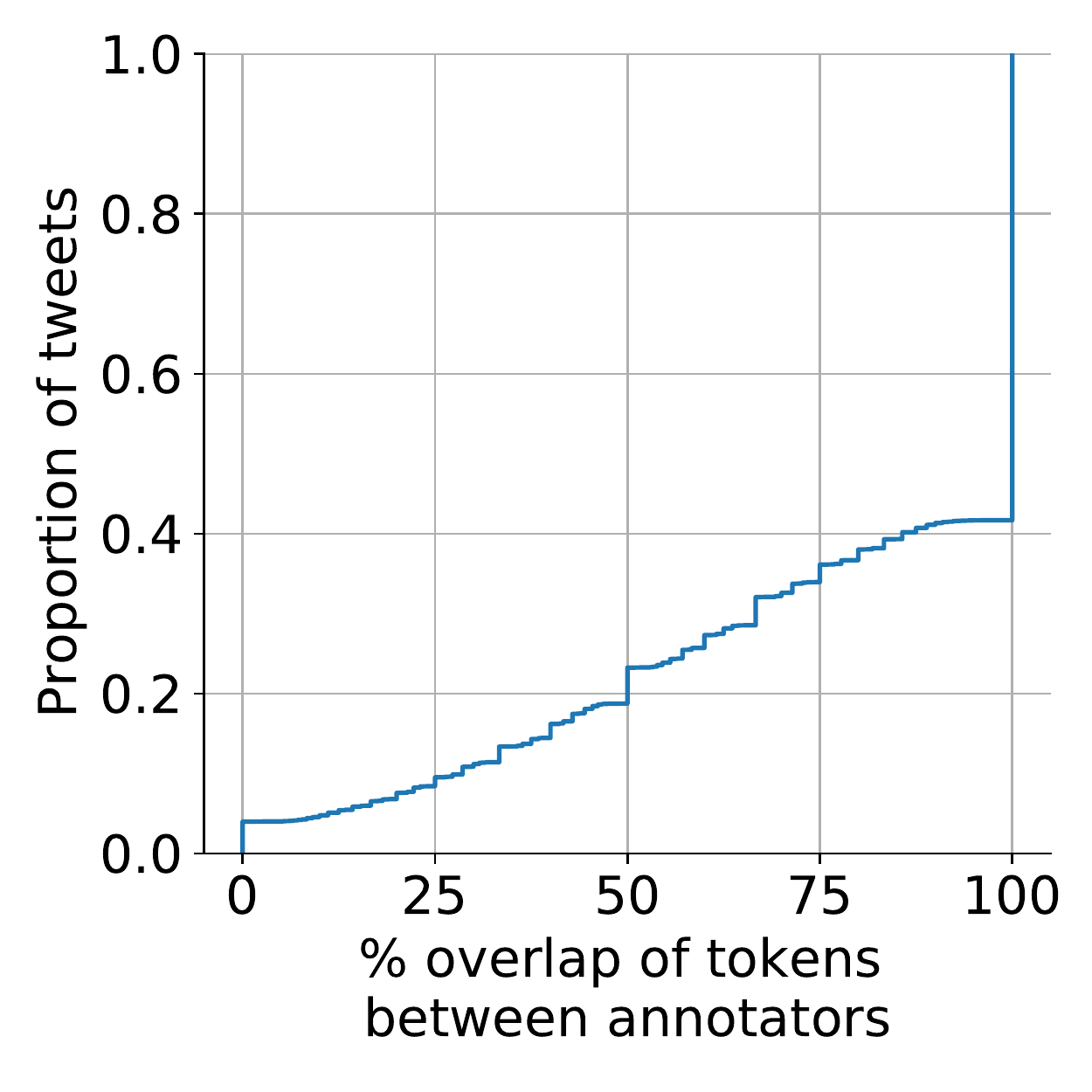}
         \caption{CDF of \% overlap between keyword-sets marked by individual annotators (intersection over union).}
         \label{fig:keywordb}
     \end{subfigure}
    % \vspace{-3mm}
    \caption{Statistics of explanations selected by the annotators}
    \label{fig:keyword_coverage}
    \Description{These figures describe the coverage of the tweet tokens by explanations and the overlap between explanations from different annotators. The first figure shows that most of the explanations do not cover a significant portion of the tweet. The second figure shows that in most of the cases the annotators agree on the keywords.}
    % \vspace{-3mm}    
\end{figure}

\begin{table}[t]
    \centering
    \small
    % \footnotesize
    \caption{Some of the most frequent keywords from the explanations for each class (after removing `\textit{covid}', `\textit{corona}', `\textit{virus}', `\textit{vaccine}' and generic stopwords).}
    \label{tab:top_keys}
    \begin{tabular}{|l|p{60mm}|}
    \hline
    \textbf{Class} & \textbf{Keywords}\\
    \hline \hline
	 Conspiracy & population, depopulation, control, world, chip, agenda, plan \\
	 \hline
	 Country & russian, chinese, china, russia, want, develop, accept \\
	 \hline
	 Ineffective & effective, efficacy, work, pfizer, get, stop, prevent \\
	 \hline
	 Ingredients & cell, aborted, chip, ingredient, fetal, tissue, contain \\
	 \hline
	 Mandatory & force, passport, mandatory, push, people, mandate, passports \\
	 \hline
	 Pharma & pfizer, pharma, money, company, gates, moderna, billion, profit \\
	 \hline
	 Political & government, trump, election, political, pfizer, borisjohnson, politician\\
	 \hline
	 Religious & catholic, religion, catholics, avoid, leader, bishop, morally \\
	 \hline
	 Untested & rush, trial, test, experimental, untested, testing, datum \\
	 \hline
	 Side-effect & die, effect, death, reaction, pfizer, clot, cause, adverse \\
	 \hline
	 Unnecessary & need, don't, take, want, people, rate, vaccinate \\
	 \hline
    \end{tabular}
    % \vspace{-8mm}    
\end{table}

Recall that for a particular tweet $t$, we consider only those labels (classes) that have been marked by at least 2 annotators.
Each annotator also marks some keywords as an explanation of each label that he/she assigns to the tweet $t$.
Now, even if two annotators label $t$ with the same label $l$, there can be differences in the explanations they give for the same label. 
Hence, for each selected label $l$ for $t$, we need to take a combination of all the keywords marked by the 2 or more annotators (as explanations) who labeled $t$ with the label $l$. 
A few examples of such explanations have been given in Table~\ref{tab:explanations_example}. 

To combine the explanations provided by multiple annotators for the same label for $t$, we can either take the \textit{union} or the \textit{intersection} of the keywords that have been marked by at least two annotators. 
After observing the explanations of about 100 tweets, we felt that taking intersection reduces the amount of information available about the explanations, making it insufficient for training  classifiers. 
%One such example is given in the first row of Table~\ref{tab:explanations}, where there is no overlap between the keywords given by the different annotators.
%On the other hand, the concern about taking the unions is that it can bring in too much useless information. However this doesn't seem to be the case. 
On the other hand, taking the union of the explanations (i.e., considering all words that have been marked as explanation by any annotator) seemed to give an appropriate amount of information. 
Hence, for label $l$ for tweet $t$, we consider as explanation the union of the keywords selected by all annotators who labeled $t$ with $l$.

We now check the quality of the explanations (that are generated by taking union of the keywords selected by multiple annotators) in various ways.
First, we wanted to check if too many tweets have a large portion covered up with explanations; if so, this could imply that the explanations do not contain enough distinct meaningful information to guide models about the important parts of the tweet.
To this end, we checked the distribution of the \% of tokens in the tweet-text that is covered by the explanations, i.e., 
the number of tokens in the explanations expressed as a percentage of number of tokens in the entire tweet.
The distribution is plotted in Figure~\ref{fig:keyworda}. 
We see that, only in very few cases ($<$1\%) does the explanations cover more than 80\% of the tweet-text; whereas, for more than 80\% of the tweets, the explanations (considering union of keywords selected by multiple annotators) cover less than 30\% of the tweet-text. 
This implies that even after taking unions, the explanations are terse enough to produce distinctive information. 
The $<$1\% cases where the explanations cover most of the tweet, seem to be cases where the tweets are small and most of it is actually relevant to the corresponding class label. 
Such an example is given in the last row of Table~\ref{tab:explanations_example}.

Second, to understand the inter-annotator agreement about explanations, we computed the distribution of the \% overlap of tokens between the keywords marked by different annotators (for the same tweet and label). 
We adopt an approach based on (intersection over union).
For each tweet we calculated the number of keywords that were marked by at least 2 annotators (intersection), and divided it by the number of keywords that were marked by at least 1 annotator (union). This distribution is given in Figure~\ref{fig:keywordb}. As can be seen from this figure, in more than 60\% of the tweets, the  annotators have complete overlap of the keywords they marked, thus implying good agreement between the annotators as well as the distinct nature of the explanations found in the tweet.
%Thus we decided to use the union of individual keywords to generate the final explanations. 

Finally, Table~\ref{tab:top_keys} shows some of the most frequent words in the final explanations associated with the different labels/classes. It is evident that the explanations are of good quality, containing keywords that are very relevant to the labels.

% This approach also has its caveats -- in some cases the explanations tend to cover up almost the entire tweets, especially when the length of the tweet is small. Such an example is given in the last row of Table~\ref{tab:explanations}.
% However, the repercussions of this is negligible in comparison to taking the intersections, due to two reasons -- i)~such cases exist when the entire tweet contains information about the particular label; and ii)~Less than 1\% tweets have explanations that cover up more than 80\% of the tweet. 
% This can be seen from Figure~\ref{fig:keyword_coverage}.
% \noteng{Is there a graph cdf??showing fraction of length of tweet vs explanation lenght}

%%%%%%%%%%%%%%%%%%%%%%%%%%%%%%%%%%%%%%%%%%%%%%%%%%%%%%%%%%%%%%%%%%%%%%%%%%
\subsection{Generating class-wise summaries}

To enable the use of the CAVES dataset for summarization, we wanted to provide summaries of the tweets from each class. 
Since the `Religious` class has only 46 tweets (as seen in Table~\ref{tab:classes}), we decided to exclude this class from the summary generation process. The `None' class is also excluded, since it does not represent any particular theme of discussion.
The labelled tweets are then segregated into 10 groups based on the remaining 10 classes -- we put a tweet into a class if the class is present in any one of its labels. 
Thus a particular tweet can be included in multiple groups if it has multiple labels.\footnote{Due to this reason, the summaries for a particular class can contain some information about other classes, due to the multi-label nature of the tweets.}

\vspace{2mm}
\noindent \textbf{Writing summaries:}
% We contacted the annotation firm once again after about 2 weeks to get some summaries generated of the labelled tweets. 
For writing the summaries of the tweets, we employed workers on the crowdsourcing platform Prolific (\url{https://prolific.co/}). 
We selected workers who are fluent in English and are conversant with Twitter. Additionally, to ensure quality annotations we also added filters selecting workers who had completed at least 1000 tasks on the platform with a 100\% acceptance rate.
% we asked the workers to provide a summary of the set of tweets in a group, and not necessarily the particular theme of the class.}. 
% Thus the same tweets can be present in multiple groups of classes. 
% We specifically asked the firm to employ different annotators this time than the ones who had worked on the previous annotation of labels and explanations \noteb{which the firm agreed}. This was done to prevent any potential biases that the previous annotators may have while dealing with the same set of tweets again. \noteb{Also the gap in time was deliberately maintained to reduce the chance of any potential slip.} \noteng{I think let us remove the time concept, is not looking very convincing.}
Each group (class) of tweets were summarized by 3 different Prolific workers. 
We asked the annotators to write \textit{abstractive} summaries. 
Specifically, the annotators were given the following instructions --
\noindent \textit{Write a coherent summary of 200 - 250 words of all the tweets you read. Write this summary in your own words. Try to cover as much of the content as you can in the summary, and also avoid redundancy as much as possible.}

We asked 3 annotators to write summaries for each class of tweets. 
Thus we finally compiled a list of 30 summaries (10 classes $\times$ 3 summaries).
% \noteb{Try to include opinions about different persons / companies / government entities.}\noteng{was this instruction necessary??}}
% \noindent \noteb{Note that we asked the annotators to provide a summary of the set of tweets in a group, and not necessarily the particular theme of the class.} \noteng{not clear} 
% We agreed to pay INR~20,000 ($\sim$USD~300) for this task and the summaries were written in around 2 more weeks.

\begin{table}[!t]
    \centering
    % \footnotesize
   % \small
    \caption{Evaluation scores of the class-wise summaries (averaged over all 3 summaries in each class).}
    \label{tab:summary_evaluation}
    \begin{tabular}{|l|c|c|c|}
    \hline
    \textbf{Class} & \textbf{Consistency} & \textbf{Fluency} & \textbf{Relevance} \\
    \hline \hline
    Conspiracy & 4.445 & 4.223 & 4.000 \\
    \hline
    Country & 4.000 & 3.556 & 3.776 \\
    \hline
    Ineffective & 3.999 & 4.332 & 3.778 \\
    \hline
    Ingredients & 4.556 & 4.334 & 4.223 \\
    \hline
    Mandatory & 3.777 & 4.334 & 4.111 \\
    \hline
    Pharma & 3.776 & 3.889 & 4.111 \\
    \hline
    Political & 3.889 & 3.667 & 4.445 \\
    \hline
    Rushed & 4.445 & 4.000 & 4.333 \\
    \hline
    Side-Effect & 3.778 & 3.889 & 4.112 \\
    \hline
    Unnecessary & 4.112 & 4.112 & 4.112 \\
    \hline \hline
    \textbf{Overall} & 4.077 & 4.033 & 4.111 \\
    \hline
    \end{tabular}
    % \vspace{-8mm}
\end{table}

\vspace{2mm}
\noindent \textbf{Evaluation of summaries:}
We next evaluate the quality of the summaries written by the Prolific workers. 
To this end, we tried three different quality metrics for summaries~\cite{fabbri2021summeval}:

\noindent $\bullet$ \textbf{Consistency} --  A consistent summary contains only statements / information that are present in the original set of tweets, and not any additional information.

\noindent  $\bullet$ \textbf{Fluency} - The quality of individual sentences in the summary with regards to how easy they are to read and understand, whether the sentences in the summary are grammatically correct, etc.

\noindent $\bullet$ \textbf{Relevance} - The summary should include only important information from the original set of tweets. Summaries which do not contain enough important information, or redundancies or excess information are given a lower score.

We again employed some Prolific workers for evaluating the summaries, taking care to choose different workers for the evaluation, and {\it not} the same workers who wrote the summaries. %but blacklisted the IDs of the workers who had written the summaries. 
Three workers were assigned to rate each of the 3 summaries individually for each of the 10 classes. 
They were also given the original set of tweets shown to the summarizing workers, as reference.
For each of the metrics defined above, the workers were asked to rate a summary on a scale of 1--5, with 1 being the worst/lowest score and 5 being the best/highest score. 

We found 4 particular summaries to be of low quality (evaluation metrics $< 3.5$). We removed these summaries and again floated the summary-writing task on Prolific, followed by evaluation of the summaries. 
All of these were done by different workers than the ones who were previously employed (which is possible to set on Prolific). 
%(by temporarily blacklisting their IDs).
The scores on the final set of summaries were averaged class-wise and have been stated in Table~\ref{tab:summary_evaluation}. 
We see that the final class-wise summaries are of sufficiently good quality, achieving scores higher than 3.5 (out of 5) for all three metrics.

%%%%%%%%%%%%%%%%%%%%%%%%%%%%%%%%%%%%%%%%%%%%%%%%%%%%%%%%%%%%%%%%%%%%%%%%%%
\subsection{CAVES dataset: overview and availability} \label{sub:availability}

\begin{table}[!t]
    \centering
    % \footnotesize
   % \small
    \caption{Overview of the CAVES dataset.}
    \label{tab:dataset_summary}
    \begin{tabular}{|l|c|}
    \hline
%    \# Tweets collected & 100M \\
%    \hline
    \# Tweets classified as Anti-Vax with high confidence & 7.5M \\
    \hline
    \# Distinct Anti-vax tweets after duplicate removal & 6.5M \\
    \hline
    \# Tweets annotated by human workers & 11,000 \\
    \hline
    \hline
    \# Labeled tweets in final dataset & 9,921 \\
    \hline
    Total \# label-explanation tuples & 10,802 \\
    \hline
    Total \# of summaries & 30\\
    \hline
    \end{tabular}
    % \vspace{-8mm}
\end{table}

The CAVES dataset contains 9,921 anti-vax tweets labeled with their anti-vaccine concerns (class), explanations for the labels, and class-wise summaries.
Table~\ref{tab:dataset_summary} gives a summary of the dataset.
For the classification and explanation generation tasks, we divide the set of tweets into train, validation, and test sets (split in the ratio 70-10-20 as defined later in Section~\ref{sec:multilabel}). 

%and the combined set of tweets in standard CSV files, containing the three labels and their corresponding explanations, as a set of cleaned words from the tweet. We also provide a script that loads this data into appropriate Numpy arrays-- i)~The tweet texts, ii)~The labels as multi-hot encoded vectors, and iii)~Explanations as matrices of multi-hot vectors per class, by applying the Longest Common Subsequence algorithm to the keywords and cleaned tweet texts.
%The summaries for each class are provided as normal text files in separate folders for each.

Note that, according to the 
Twitter sharing policy\footnote{\url{https://developer.twitter.com/en/developer-terms/more-on-restricted-use-cases}}, it is only permitted to share the tweet IDs, whereas the tweet text (or hydrated tweet objects) should not be shared publicly. 
In compliance with the Twitter policy, we have publicly released the tweet-IDs, the labels and the abstractive summaries on Github at \url{https://github.com/sohampoddar26/caves-data}. 
We will be providing the full-text and the explanations on request to any researcher who agrees to use the dataset for research purposes.
Detailed instructions are given in the Github repository.

We have also made publicly available the implementations of the benchmark methods tried on the dataset (as described in the next section) as well as some helper scripts (such as scripts for loading the dataset into appropriate Numpy arrays).

% For the purpose of review, we have uploaded a small data sample in a google drive folder~\footnote{\url{https://drive.google.com/drive/folders/1TRRGHrIW634nFMyVgkT2bFFbNH0x_opd}} containing all the tweet-IDs in the final labeled dataset, some sample tweets from each class along with their labels and explanations, as well as their summaries. 
% \footnote{\url{https://tinyurl.com/sigir22poddar}}

\section{Tasks on the Dataset}
\label{sec:tasks}

In this section we shall discuss three tasks that can be directly applied to the dataset -- (1)~multi-label classification, (2)~explainable classification, and (3)~tweet summarization. 
We also apply several state-of-the-art methods for benchmarking each of the three tasks.

\subsection{Multi-label classification}
\label{sec:multilabel}

In the standard multi-label classification task, each data point (tweet in our case) has to be assigned to one or more classes (anti-vaccine concerns in our case). 
This is in contrast with the multi-class (or single-label) classification in which a data point has to be assigned with exactly one (out of many) classes.

\vspace{1.5mm}
\noindent \textbf{Data splitting:} 
For creating a benchmark, we split the tweet-set into a train, a validation and a test set in the ratio of 70-10-20. 
Since the distribution of classes is skewed (as seen in Table~\ref{tab:classes}), a random split would cause a class imbalance. Thus we decided to split the data using the iterative stratification method~\cite{sechidis2011stratification,szymanski2017network} that aims to balance the classes in a multi-label setting. We used the sk-multilearn library~\cite{szymanski2017scikit} to perform this stratification.

\vspace{1.5mm}
\noindent \textbf{Metrics for Evaluation:}
Given the set of predicted and gold standard labels, we have used 4 different metrics to estimate the performance of the models. 
First we calculate the F1-score for all the 12 classes separately and find the (i)~Macro-average, and (ii)~Weighted-average (where the weights of the classes are proportional to the class frequencies).
We also calculate the Jaccard similarity between the predicted label-set and the gold standard label-set over each tweet, and average the Jaccard similarity values over all tweets. 
Finally we also calculate the subset accuracy -- for a particular tweet, a predicted set of labels is considered a match only if it \textit{exactly} matches with the set of gold standard labels.
All metrics were calculated using standard functions from the Scikit-Learn package~\cite{scikit-learn}.

\begin{table}[tb]
    \centering
    % \footnotesize
    \small
    \caption{Comparison of models for multi-label classification.}
    \label{tab:multi-label-results}
    \begin{tabular}{|p{14mm}|c|c|c|c|}
    \hline
        \textbf{Model} & \textbf{Macro-F1} & \textbf{Weighted-F1} & \textbf{Jaccard} & \textbf{Accuracy} \\
        \hline \hline
        \multicolumn{5}{|l|}{\textit{Only predicts labels}}\\
        \hline
        LLDA & 0.1188 & 0.2134 & 0.1543 & 0.0675 \\
        \hline \hline
        RoBERTa & 0.6626 & 0.7319 & 0.6949 & 0.6004 \\
        \hline
        CT-BERT & 0.6924 & 0.7419 & 0.7040 & 0.6054 \\
        \hline
        HateXplain & 0.6709 & 0.7245 & 0.6909 & 0.5906 \\
        % \hline
        % \todo{EFL} &  &  &  &  \\
        \hline \hline
        \multicolumn{5}{|l|}{\textit{Predicts labels and explanations}}\\
        \hline
        CAML & 0.4286 & 0.5341 & 0.4562 & 0.3530 \\
        \hline
        ExPred & 0.6558 & 0.7105 & 0.6576 & 0.5726 \\
        \hline
        Multi-Task & 0.6823 & 0.7371 & 0.7018 & 0.6004 \\
        \hline
        Multi-Task (w GRU) & 0.6779 & 0.7424 & 0.7080 & 0.5989 \\
    \hline
    \end{tabular}
\end{table}

% \begin{table}[tb]
%     \centering
%     % \footnotesize
%   % \small
%     \caption{Comparison of models for multi-label classification.}
%     \label{tab:multi-label-results}
%     \begin{tabular}{|p{14mm}|c|c|c|c|}
%     \hline
%         \textbf{Model} & \textbf{Macro-F1} & \textbf{Weighted-F1} & \textbf{Jaccard} & \textbf{Accuracy} \\
%         \hline \hline
%         \multicolumn{5}{|l|}{\textit{Only predicts labels}}\\
%         \hline
%         LLDA & 0.1188 & 0.2134 & 0.1543 & 0.0675 \\
%         \hline \hline
%         RoBERTa & 0.5837 & 0.6993 & 0.6713 & 0.6010 \\
%         \hline
%         CT-BERT & 0.5860 & 0.7144 & 0.6884 & 0.6087 \\
%         \hline
%         HateXplain & \textbf{0.6007} & \textbf{0.7153} & \textbf{0.6928} &  \textbf{0.6294} \\
%         % \hline
%         % \todo{EFL} &  &  &  &  \\
%         \hline \hline
%         \multicolumn{5}{|l|}{\textit{Predicts labels and explanations}}\\
%         \hline
%         CAML & 0.3449 & 0.5039 & 0.4365 & 0.3675 \\
%         \hline
% %        Explain then predict & 0.5663 & 0.6867 & 0.6474 & 0.5675 \\
% %        \hline
%         ExPred & 0.5870 & 0.6924 & 0.6652 & 0.5715 \\
%         \hline
%         Multi-Task & 0.5775 & 0.7029 & 0.6761 & 0.5937 \\
%         \hline
%         Multi-Task (w GRU) & \textbf{0.5991} & \textbf{0.7104} & \textbf{0.6855} & \textbf{0.6037} \\
%     \hline
%     \end{tabular}
% \end{table}

%%%%%%%%%%%%%%%%%%%%%%%%%%%%%%%%%%%%%%%%%%%%%%%%%%%%%%%%%%%%%%%%%%%%
% \subsection{Baselines}
\vspace{1.5mm}
\noindent \textbf{Benchmarking methods:}
We experimented with representative methods from different families of multi-label classifiers.
We try out a topic modelling method (Labeled LDA) since these are what have been used by prior works for understanding anti-vaccine opinions (as described in Section~\ref{sec:relwork_concerns}).
We tried some basic transformer based classifiers, namely RoBERTa-Large (known to perform very well for several NLP tasks) and CT-BERT-v2~\cite{muller2020covid} which is a BERT-Large model pretrained on 
% several million tweets related to 
millions of COVID-19 tweets (known to give good classification performance on COVID-related tweets~\cite{poddar2022winds}). 
Next we modified the HateXplain~\cite{mathew2021hatexplain} model to work for multi-label scenario by joining the explanations for different labels. 

We also tried a few models which incorporate the explanations, either to predict only the labels or to predict the labels and generate explanations jointly. 
In this family of models, we tried the CAML~\cite{mullenbach2018explainable} model which uses CNN and Attention mechanism and can be used to get explanations for all the labels.
We also use a simplified version of the recently-developed ExPred~\cite{zhang2021explain} model, and modify it for a multi-label setting. 
This model generates explanations along with a label prediction auxiliary task and then predicts the labels again.
%We then try some generic models-- ``\textit{Explain then predict}'' model first generates the explanations from the CT-BERT embeddings and then uses them to predict the labels. 
The ``\textit{Multi-task}'' model contains a shared CT-BERT encoder and two decoders which separately predict the labels and generate the explanations. We also try a variation of the Multi-Task model where we pass the token embeddings through a GRU first before generating the explanations.

For each of the above-stated methods, we used a batch size of 16 and a maximum sequence length of 128. 
For HateXplain in particular we obtained best results with a batch size of 2.
The models were trained and validated on corresponding datasets for 20 epochs. The model that yielded the best Macro-F1 score on the validation dataset was saved, and this model was used to calculate the performance metrics on the test dataset.

\vspace{1.5mm}
\noindent \textbf{Performances:}
The results of different methods are given in Table~\ref{tab:multi-label-results}. It is seen that CT-BERT outperforms RoBERTa slightly due to the domain pretraining. In contrast, the LLDA model vastly under-performs, highlighting the limitations of the previous works which tried to use topic modelling for extracting concerns about vaccines.
CAML does not perform too well because it uses word2vec embeddings which are known to not be as good as BERT embeddings.
The CT-BERT model seems to perform the best, achieving a Macro-F1 score of 0.6924, while the Multi-Task model  achieves a Macro-F1 score of 0.6823, while also generating explanations.
In terms of class-wise F1 score, all the models seem to perform badly on the `None' and `Conspiracy' classes which  contain some confusing arguments. The `Country' and `Religious' classes are also very sparse, which lead to models under-performing on these classes. Details are omitted for brevity.

\subsection{Explainable classification}
\label{sec:explanation}

The next task is that of generating explanations along with the predicted labels. The explanation generation part is a sequence-labelling task, where for a given tweet text as input, each token in the text is to be marked if it is part of the explanation or not. This is to be repeated for each of the predicted labels for the tweet. Thus, for a given tweet, the output of the task will be (i)~the predicted labels, and (ii)~a list of binary vectors, one for each predicted label (where each vector is of the same length as the number of tokens in the tweet). In these binary vectors, an element being 1 implies that the corresponding token in the tweet-text is part of the explanation for the corresponding label.  
% We also define a tuple-prediction task, where the model will be evaluated based on pair of predicted labels and corresponding generated explanations.
% \todo{elaborate: input is tweet text, only one task, tuple prediction isn't task, just evaluation}

% \begin{table}[!ht]
%     \centering
%     \small
%     \begin{tabular}{|l|c|c|c|c|}
%     \hline
%     \textbf{Model} & \textbf{IOU-F1} & \textbf{Jaccard} & \textbf{Token F1} & \\
%     \hline \hline
%         CAML & 0.0774 & 0.1899 & 0.2886 &  \\
%         \hline
%         Explain then predict & 0.0399 & 0.1697 & 0.2673 &  \\
%         \hline
%         ExPred & 0.0560 & 0.1582 & 0.2487 & \\
%         \hline
%         Multi-Task & 0.3476 & 0.3336 & 0.4154 & \\
%         \hline
%         Multi-Task (w GRU) & 0.3015 & 0.2920 & 0.3451 & \\
%     \hline
%     \end{tabular}
%     \caption{Performance metrics of different models on explanation generation task}
%     \label{tab:explanations}
% \end{table}

\begin{table}[!t]
    \centering
    % \footnotesize
    \small
    \caption{Comparison of models for explanation generation.}
    \label{tab:explanations}
    \begin{tabular}{|p{15mm}|c|c|c|}
    \hline
    \textbf{Model} & \textbf{Tuple-Pre} & \textbf{Tuple-Rec} & \textbf{Tuple-F1} \\
    \hline \hline
        CAML & 0.0099 & 0.0082 & 0.0089 \\
        \hline
        ExPred & 0.1944 & 0.1535 & 0.1716 \\
        \hline
        Multi-Task & \textbf{0.3952} & \textbf{0.3961} & \textbf{0.3957} \\
        \hline
        Multi-Task (with GRU) & 0.3304 & 0.3383 & 0.3343 \\
    \hline
    \end{tabular}
    % \vspace{-8mm}
\end{table}

% \begin{table}[!t]
%     \centering
%     % \footnotesize
%   % \small
%     \caption{Comparison of models for explanation generation.}
%     \label{tab:explanations}
%     \begin{tabular}{|p{15mm}|c|c|c||c|c|}
%     \hline
%     & \multicolumn{3}{|c||}{\it Explanation metrics} & \multicolumn{2}{c|}{\it Tuple metrics} \\ 
%     \hline
%     \textbf{Model} & \textbf{IOU-F1} & \textbf{Jaccard} & \textbf{Token F1} & \textbf{M-F1} & \textbf{Acc} \\
%     \hline \hline
%         CAML & 0.0774 & 0.1899 & 0.2886 & 0.0351 & 0.0498 \\
%         \hline
%         % Explain then predict & 0.0399 & 0.1697 & 0.2673 & 0.0117 & 0.0101  \\
%         % \hline
%         ExPred & 0.0560 & 0.1582 & 0.2487 & 0.0249 & 0.0156 \\
%         \hline
%         Multi-Task & 0.3476 & 0.3336 & 0.4154 & 0.2704 & 0.2854 \\
%         \hline
%         Multi-Task (with GRU) & 0.3015 & 0.2920 & 0.3451 & 0.2355 & 0.2724 \\
%     \hline
%     \end{tabular}
%     % \vspace{-8mm}
% \end{table}

% \begin{table}[!ht]
%     \centering
%     \small
%     \begin{tabular}{|l|c|c|c|c|}
%     \hline
%     \textbf{Model} & \textbf{Macro-F1} & \textbf{Accuracy} \\
%     \hline \hline
%         CAML & 0.0351 & 0.0498 \\
%         \hline
%         Explain then predict & 0.0117 & 0.0101  \\
%         \hline
%         ExPred & 0.0249 & 0.0156 \\
%         \hline
%         Multi-Task & 0.2704 & 0.2854 \\
%         \hline
%         Multi-Task (w GRU) & 0.2355 & 0.2724 \\
%     \hline
%     \end{tabular}
%     \caption{Performance metrics of different models on tuple extraction task}
%     \label{tab:tuples}
% \end{table}

\begin{table}[!t]
    \centering
    % \footnotesize
    \small
    % \resizebox{\columnwidth}{!}{
    \caption{Comparison of models for Summarization task.}
    \label{tab:summary_results}
    \begin{tabular}{|l|c|c|c|}
        \hline
        \textbf{Model} & \textbf{ROUGE-1 F1} & \textbf{ROUGE-2 F1} & \textbf{ROUGE-L F1} \\
        \hline \hline
        \multicolumn{4}{|l|}{\textbf{Extractive models}} \\
        \hline
        LexRank~\cite{erkan2004lexrank} & 21.48 & 2.29 & 15.73 \\
        \hline
        PacSum~\cite{zheng2019sentence} & 19.14 & 2.09 & 12.69 \\
        \hline
        COWTS~\cite{rudra2015extracting} & 31.54 & 4.52 & 21.07\\
        \hline \hline
        \multicolumn{4}{|l|}{\textbf{Abstractive models}} \\
        \hline
        BART~\cite{lewis2020bart} & 28.60 & 5.04 & 19.19 \\
        \hline
        Pegasus~\cite{zhang2020pegasus} & 25.20 & 5.16 & 20.00 \\
        \hline
        T5~\cite{raffel2019exploring} & 27.73 & 5.34 & 19.38\\
        \hline
    \end{tabular}
    % }
    % \vspace{-8mm}
\end{table}

\vspace{1.5mm}
\noindent \textbf{Metrics for evaluation:}
For evaluating the explanations generated by different methods, we use some of the hard selection metrics defined by~\citet{deyoung2020eraser}.
Among these, the first metric IOU-F1 (Intersection-Over-Union F1) is useful for detecting partial matches. 
The IOU is the size of the overlap between two sets of tokens divided by the size of the union of the two sets of tokens (also known as the Jaccard coefficient). 
For a particular tweet and label, a predicted explanation is counted as a match if and only if the IOU with the ground truth explanation is $\geq 0.5$~\cite{deyoung2020eraser}. 
% These matches are then used to calculate the multi-label F1 score as described previously in Section~\ref{sec:multilabel}. 
% Additionally, the token-level F1 score and Jaccard coefficient is also calculated and averaged over all data points. 
% We collectively refer to these metrics as {\it Explanation metrics}.

% Separately, we also 
We then \textit{evaluate the tuples of predicted labels and explanations together}. 
We consider a (label, explanation) tuple to be a match if and only if the predicted label is present in the gold standard set of labels and the predicted explanation overlaps at least 50\% with the corresponding gold standard explanation (IOU $\geq 0.5$). The Precision, Recall and F1-score is then calculated over these matches as before. We refer to these metrics as {\it Tuple metrics}. 
Note that these classification metrics cannot be calculated for the ``non-matching'' class, since we consider a match if there is sufficient overlap in the explanations (and the labels are same), and thus a non-match is not easy to define. Hence the macro-F1 scores cannot be calculated either.

\vspace{1.5mm}
\noindent \textbf{Benchmarking methods:}
For this task, we use the same models which generate explanations (CAML, ExPred, Multi-Task and Multi-Task w GRU), as described in the previous subsection on multi-label classification. We also use the same data splits along with the experimental settings. Hence we are omitting these details here for brevity. It must however be noted that for the `CAML' model, the explanations were extracted from the Attention layer weights, by taking the top-8 tokens with the highest weights. For the rest of the models, the explanations were predicted as a sequence-labelling task, with those tokens being generated that have the sigmoid of the logits $\geq 0.5$ (similar to a multi-label prediction).

\vspace{1.5mm}
\noindent \textbf{Performances:}
The metric scores on the test set for the explanation generation task are given in Table~\ref{tab:explanations}.
% , while those for the tuple-prediction task is given in Table~\ref{tab:tuples}.
The two variations of the  Multi-task models seem to perform the best on these tasks, with the standard model achieving an Tuple-F1 score of 0.3957 and the one with a GRU achieving 0.3343. 
% Especially for the tuple metrics, these models 
%achieve Macro-F1 scores of 0.2704 and 0.2355 respectively.
These models seem to comprehensively outperform the CAML and ExPred models for the explainable classification task.
% Similar to the multi-class classification scores, the IOU-F1 scores are low for the `Country' and `Religious' (sparse) classes.
%For the best Multi-Task model, the IOU-F1 score of explanations were particularly less for the political class.

%\noteng{I think some error analysis commenting on the common mistakes the models are doing would be good. }

\subsection{Summarization}
\label{sec:summarization}

Multi-document summarization (a special case of which is tweet summarization) is a classical IR task that aims to produce a short summary of a large set of documents, that contains as much information content as possible while reducing redundant information. Summarization algorithms come in two flavours -- i)~Extractive, which select a few tweets out of all the tweets, and ii)~Abstractive, which generates words to create a coherent summary like humans do. 
We have tried a few methods of each type to generate class-wise summaries and get some benchmark scores. 
% We specially attempt .

\vspace{1.5mm}
\noindent \textbf{Metrics for Evaluation:}
For evaluation of the summaries, we use the popular ROUGE metrics. Specifically, we report the ROUGE-1 F1-score (that considers unigram matches between the gold standard summaries and an algorithm-generated summary), ROUGE-2 F1-score (that considers bigram matches) and the ROUGE-L F1-score (that considers longest sequence matches). 
%ROUGE-1 Recall is calculated as the fraction of unigrams in the ground truth summary that are present in the predicted summary too, while the ROUGE-1 Precision is the fraction of unigrams in the ground truth summary that are present in the predicted summary. ROUGE-1 F1-score is the harmonic mean of the ROUGE-1 Recall and ROUGE-1 Precision. 
%Similarly, ROUGE-2 and ROUGE-L F1-scores are calculated while considering the bigrams; and the lengths of longest sequence matches instead of just the unigrams.

\vspace{1.5mm}
\noindent \textbf{Benchmarking methods:}
We have employed a few popular summarization algorithms to benchmark our summarization dataset. 
Among extractive methods, we have used the graph-based LexRank summarizer~\cite{erkan2004lexrank}, 
%which is a variation of the TextRank algorithm 
PacSum~\cite{zheng2019sentence} that defines its own centrality measures, and COWTS~\cite{rudra2015extracting} which is an Integer Linear Programming based summarizer designed for disaster-related tweets.
Among abstractive methods, we have employed different {\it pretrained} transformer based encoder-decoder models such as T5~\cite{raffel2019exploring},  BART~\cite{lewis2020bart}, and Pegasus~\cite{zhang2020pegasus}, that differ in their pre-training strategies.

For each class, we used the extractive models to generate summaries of around 20 tweets each, from the document consisting of all labeled tweets belonging to that class.
% For the abstractive models we used them to generate summaries of up to 250 words. 
Similarly, the abstractive models were used to generate class-wise summaries of around 250 words each.
It is to be noted here that the pre-trained abstractive models have a limitation on the size of input documents that they can summarize.
Hence, for each class, we first split the corresponding tweet dataset into almost equal-sized chunks such that the length of each chunk is less than or equal to the maximum permissible length.
We obtain smaller summaries from each chunk and concatenate these to form a longer summary.
The final summary of length around 250 words was obtained by greedily selecting the top-ranked sentences (from the longer summary) based on their TF-IDF scores w.r.t. all tweets in the corresponding class.

\vspace{1.5mm}
\noindent \textbf{Performances:}
The ROUGE scores on the summarization task are given in Table~\ref{tab:summary_results}.
Among the extractive models, COWTS performs the best achieving a ROUGE-2 F1 score of 4.52 and a ROUGE-L F1 score of 21.07.
The performances of the different abstractive models are similar. 
%T5 achieves the best ROUGE-2 score of 5.34, while Pegasus achieves the best ROUGE-L score among the abstractive models.
All models achieve especially low ROUGE-2 F1 scores, and it is a potential research direction to improve summarization performance over this dataset.

\subsection{Summary of the section}

We tried different state-of-the-art models on three tasks to establish benchmark results on the CAVES dataset. 
For the classification task the highest Macro-F1 score achieved was 0.6007, which is a moderate score considering it is a multi-label setting. 
Explanations generated were of decent quality with the best IOU-F1 of 0.3476, though the performance for some other models seem to be quite low. 
For the summarization task, the ROUGE-2 scores achieved were low. This may be due to various reasons such as the specific vocabulary used in the anti-vax text may be unique, repetition of similar concepts in multiple tweets, etc. 
%hence careful reconsideration of supervised and/or tweet-specific approaches need to be undertaken.
These results  suggest that the CAVES dataset  and associated tasks pose interesting challenges,  and more specific models need to be developed to tackle the tasks.
\section{Other potential uses of the dataset}
\label{sec:extensions}

In this section, we highlight some potential applications of our proposed dataset, other than the ones we have discussed so far. 
% the multi-label classification, explanation generation and summarization.

%%%%%%%%%%%%%%%%%%%%%%%%%%%%%%%%%%%%%%%%%%%%%%%%%%%%%%%%%%%%%%%%%%%%%%%%%%%%%%%%%%%

\begin{figure}[!t]
    \centering
    \includegraphics[width=\linewidth]{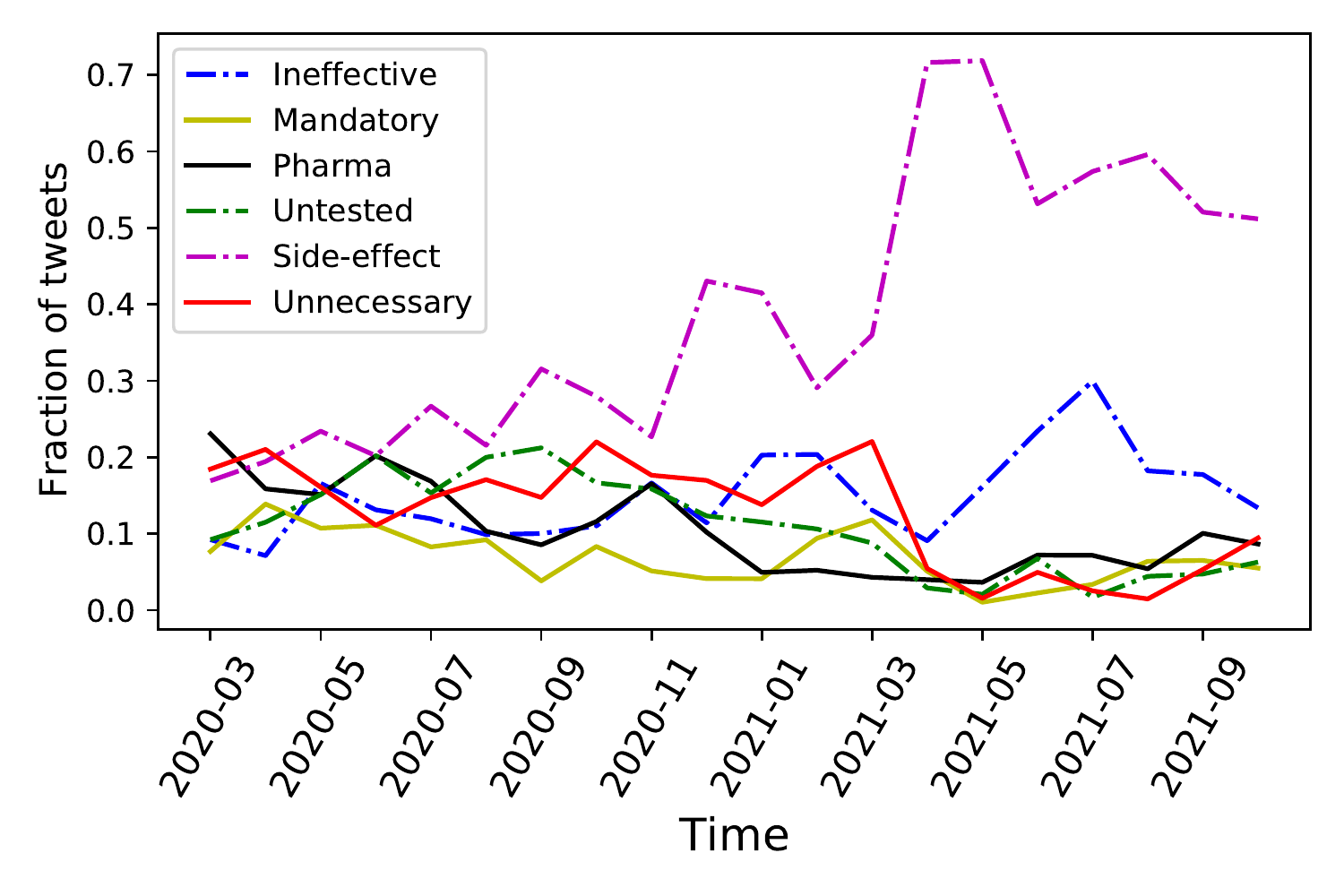}
    \caption{Frequency distribution of tweets corresponding to different concerns over time.}
    \label{fig:label_time}
    \Description{This figure shows the temporal variation of different classes in the dataset. Some classes have peaks and valleys at certain times which can correlate with real world events.}
    % \vspace{-6mm}
\end{figure}

\vspace{2mm}
\noindent \textbf{Distribution of concerns over time:}
Our dataset can be used to study the change in the distribution of different anti-vaccine concerns over time (and for training models to track such changes when applied on large scale Twitter data). 
%This can be especially helpful in understanding the changing dynamics of concerns that people might have over a period of time, and possibly mapping them to specific real-world events.
% We have simulated this with just the labelled tweets from the dataset, and plotted the month-wise distribution of the top 6 classes in Figure~\ref{fig:label_time}.
In Figure~\ref{fig:label_time}, we show the month-wise frequency distribution of the largest 6 classes in our labeled set of tweets (in terms of the number of tweets in a class).
%using only the set of labeled tweets in our dataset. 
We observe some interesting peaks in the figure which can be mapped to certain real-world events (e.g., as reported in the AJMC articles on COVID vaccine developments throughout 2020~\cite{ajmc2020timeline}
% \footnote{\url{https://tinyurl.com/AJMC-COVID-timeline-2020}} 
and 2021~\cite{ajmc2021timeline})--
% \footnote{\url{https://tinyurl.com/AJMC-COVID-timeline-2021}})--

\noindent $\bullet$ There is a spike in the \textit{Pharma} class around September 2020 which could be explained by two events -- Pfizer expanding phase 3 trials of its vaccine, and Astrazeneca trial halting due to complications faced by a patient.

\noindent $\bullet$ There is a spike in the \textit{Unnecessary} class in October 2020, which may be due to the FDA's approval of Remdesivir as a COVID-19 drug (which made many people feel that vaccines are unnecessary).

\noindent $\bullet$ The \textit{Rushed} class has a spike in November 2020 likely due to Pfizer and AstraZeneca reporting completion of their trials.

\noindent $\bullet$ The \textit{Side-effects} class has the peak in April-May of 2021, likely due to some adverse reactions to the Johnson \& Johnson vaccine being reported at the end of April.

\noindent $\bullet$ The spike in the \textit{Ineffective} class in July 2021 is likely owing to the reports of the Pfizer Vaccine not being effective against the delta variant of COVID-19.

It is to be noted here that the labeled set of tweets in the CAVES dataset might not be fully representative of the actual temporal distribution of tweets, due to inadvertent biases that might have crept in as part of the various steps we took for selecting the tweets.
%since we had selected anti-vax tweets with high precision (as discussed in Section~\ref{sec:tweet_selection}). 
Hence, one should be careful in drawing strong trends or temporal conclusions just by analyzing the labeled dataset.
%The dataset should rather be used to train deep learning models and then apply them on some extensive set of tweets to draw representative conclusions. 

\vspace{2mm}
\noindent \textbf{Generating highlights for explainable summarization:}
The benchmark methods we have used in Section~\ref{sec:summarization} are standard summarization models that work only with the tweet texts. Our dataset can facilitate designing of models that utilize the explanations to improve the summarization task. Similar to~\citet{Wang_Gao_Bai_Lapata_Huang_2021}, a select-then-generate framework could be designed that first highlights the reasoning spans as explanations and then generates a summary while focusing on the highlighted spans. Such as approach can not only improve the interpretability of extractive summarization models, but can also provide suitable explanations behind generation of particular phrases in case of abstractive models.

%%%%%%%%%%%%%%%%%%%%%%%%%%%%%%%%%%%%%%%%%%%%%%%%%%%%%%%%%%%%%%%%%%%%%%%%%%%%%%%%%%%
\vspace{2mm}
\noindent \textbf{Conspiracy detection:}
Our dataset contains a set of tweets related to conspiracies around COVID-19 vaccines, as reported in Table~\ref{tab:classes}.
The dataset can thus be used to benchmark automatic COVID-19 conspiracy theory detection models such as~\cite{shahsavari2020conspiracy}. Our dataset can further facilitate the design and evaluation of multi-task models that not only detect conspiracy-related tweets but also generate explanations for the same.

% The corresponding explanations can also be used to build or test models that provide explanations for conspiracy detection.

%%%%%%%%%%%%%%%%%%%%%%%%%%%%%%%%%%%%%%%%%%%%%%%%%%%%%%%%%%%%%%%%%%%%%%%%%%%%%%%%%%%

\section{Conclusion}
\label{sec:conc}

We have built a dataset of tweets that is important from a societal standpoint as it identifies concerns that people have towards vaccines, as well as facilitates explainable classification in a multi-label setting. 
The dataset also contains summaries of different classes and hence can be used to develop or test summarization algorithms. 
We have provided some benchmark results on the three different primary tasks, and discussed some other potential retrieval tasks.

The benchmark results point towards the need for improved, customized models for addressing the tasks. 
For example, apart from the tweet texts, the models can potentially %the \vspace{2mm} \noindent \textbf{Future Work:}
% \todo{how to extent the dataset??}
%For the different tasks some purpose-built models should be tried to improve scores. One can also try to 
incorporate additional (meta) information from the tweets or the users who posted the tweets to improve scores. 
Given the timely importance of the CAVES dataset, we believe it will instill enough interest within the community in the near future, to develop better methods for the  proposed tasks. 
%Models that better handle imbalance in classes can also be used to improve classification scores. The explanation generation task can also be tackled as a ranking problem instead of the sequence labelling that we have tried.  For summarization one can try tweet based summarization models, or experimenting with supervised models.

% 
%%
%% The acknowledgments section is defined using the "acks" environment
%% (and NOT an unnumbered section). This ensures the proper
%% identification of the section in the article metadata, and the
%% consistent spelling of the heading.
\begin{acks}
The project is partially supported by research grants from Accenture Corporation and DRDO, Government of India (through the research project titled ``Claim Detection and Verification using Deep NLP: an Indian Perspective'').
S. Poddar is also supported by the Prime Minister's Research Fellowship (PMRF) from the Ministry of Education, Government of India.
\end{acks}

%%
%% The next two lines define the bibliography style to be used, and
%% the bibliography file.
% \newpage
\bibliographystyle{ACM-Reference-Format}
\bibliography{ref}

%%% -*-BibTeX-*-
%%% Do NOT edit. File created by BibTeX with style
%%% ACM-Reference-Format-Journals [18-Jan-2012].

\begin{thebibliography}{46}

%%% ====================================================================
%%% NOTE TO THE USER: you can override these defaults by providing
%%% customized versions of any of these macros before the \bibliography
%%% command.  Each of them MUST provide its own final punctuation,
%%% except for \shownote{}, \showDOI{}, and \showURL{}.  The latter two
%%% do not use final punctuation, in order to avoid confusing it with
%%% the Web address.
%%%
%%% To suppress output of a particular field, define its macro to expand
%%% to an empty string, or better, \unskip, like this:
%%%
%%% \newcommand{\showDOI}[1]{\unskip}   % LaTeX syntax
%%%
%%% \def \showDOI #1{\unskip}           % plain TeX syntax
%%%
%%% ====================================================================

\ifx \showCODEN    \undefined \def \showCODEN     #1{\unskip}     \fi
\ifx \showDOI      \undefined \def \showDOI       #1{#1}\fi
\ifx \showISBNx    \undefined \def \showISBNx     #1{\unskip}     \fi
\ifx \showISBNxiii \undefined \def \showISBNxiii  #1{\unskip}     \fi
\ifx \showISSN     \undefined \def \showISSN      #1{\unskip}     \fi
\ifx \showLCCN     \undefined \def \showLCCN      #1{\unskip}     \fi
\ifx \shownote     \undefined \def \shownote      #1{#1}          \fi
\ifx \showarticletitle \undefined \def \showarticletitle #1{#1}   \fi
\ifx \showURL      \undefined \def \showURL       {\relax}        \fi
% The following commands are used for tagged output and should be
% invisible to TeX
\providecommand\bibfield[2]{#2}
\providecommand\bibinfo[2]{#2}
\providecommand\natexlab[1]{#1}
\providecommand\showeprint[2][]{arXiv:#2}

\bibitem[Bonnevie et~al\mbox{.}(2021)]%
        {bonnevie2021quantifying}
\bibfield{author}{\bibinfo{person}{Erika Bonnevie}, \bibinfo{person}{Allison
  Gallegos-Jeffrey}, \bibinfo{person}{Jaclyn Goldbarg}, \bibinfo{person}{Brian
  Byrd}, {and} \bibinfo{person}{Joseph Smyser}.}
  \bibinfo{year}{2021}\natexlab{}.
\newblock \showarticletitle{Quantifying the rise of vaccine opposition on
  Twitter during the COVID-19 pandemic}.
\newblock \bibinfo{journal}{\emph{Journal of communication in healthcare}}
  \bibinfo{volume}{14}, \bibinfo{number}{1} (\bibinfo{year}{2021}),
  \bibinfo{pages}{12--19}.
\newblock


\bibitem[Cao et~al\mbox{.}(2016)]%
        {cao2016tgsum}
\bibfield{author}{\bibinfo{person}{Ziqiang Cao}, \bibinfo{person}{Chengyao
  Chen}, \bibinfo{person}{Wenjie Li}, \bibinfo{person}{Sujian Li},
  \bibinfo{person}{Furu Wei}, {and} \bibinfo{person}{Ming Zhou}.}
  \bibinfo{year}{2016}\natexlab{}.
\newblock \showarticletitle{Tgsum: Build tweet guided multi-document
  summarization dataset}. In \bibinfo{booktitle}{\emph{Proceedings of the AAAI
  Conference on Artificial Intelligence}}, Vol.~\bibinfo{volume}{30}.
\newblock


\bibitem[Cotfas et~al\mbox{.}(2021)]%
        {cotfas2021longest}
\bibfield{author}{\bibinfo{person}{Liviu-Adrian Cotfas},
  \bibinfo{person}{Camelia Delcea}, \bibinfo{person}{Ioan Roxin},
  \bibinfo{person}{Corina Ioan{\u{a}}{\c{s}}}, \bibinfo{person}{Dana~Simona
  Gherai}, {and} \bibinfo{person}{Federico Tajariol}.}
  \bibinfo{year}{2021}\natexlab{}.
\newblock \showarticletitle{The longest month: Analyzing covid-19 vaccination
  opinions dynamics from tweets in the month following the first vaccine
  announcement}.
\newblock \bibinfo{journal}{\emph{IEEE Access}}  \bibinfo{volume}{9}
  (\bibinfo{year}{2021}), \bibinfo{pages}{33203--33223}.
\newblock


\bibitem[DeYoung et~al\mbox{.}(2020)]%
        {deyoung2020eraser}
\bibfield{author}{\bibinfo{person}{Jay DeYoung}, \bibinfo{person}{Sarthak
  Jain}, \bibinfo{person}{Nazneen~Fatema Rajani}, \bibinfo{person}{Eric
  Lehman}, \bibinfo{person}{Caiming Xiong}, \bibinfo{person}{Richard Socher},
  {and} \bibinfo{person}{Byron~C Wallace}.} \bibinfo{year}{2020}\natexlab{}.
\newblock \showarticletitle{ERASER: A Benchmark to Evaluate Rationalized NLP
  Models}. In \bibinfo{booktitle}{\emph{Proceedings of the 58th Annual Meeting
  of the Association for Computational Linguistics}}.
  \bibinfo{pages}{4443--4458}.
\newblock


\bibitem[Dhama et~al\mbox{.}(2021)]%
        {dhama2021covid}
\bibfield{author}{\bibinfo{person}{Kuldeep Dhama}, \bibinfo{person}{Khan
  Sharun}, \bibinfo{person}{Ruchi Tiwari}, \bibinfo{person}{Manish Dhawan},
  \bibinfo{person}{Talha~Bin Emran}, \bibinfo{person}{Ali~A Rabaan}, {and}
  \bibinfo{person}{Saad Alhumaid}.} \bibinfo{year}{2021}\natexlab{}.
\newblock \showarticletitle{COVID-19 vaccine hesitancy--reasons and solutions
  to achieve a successful global vaccination campaign to tackle the ongoing
  pandemic}.
\newblock \bibinfo{journal}{\emph{Human Vaccines \& Immunotherapeutics}}
  \bibinfo{volume}{17}, \bibinfo{number}{10} (\bibinfo{year}{2021}),
  \bibinfo{pages}{3495--3499}.
\newblock


\bibitem[Dutta et~al\mbox{.}(2018)]%
        {dutta-ensemble-summarization}
\bibfield{author}{\bibinfo{person}{Soumi Dutta}, \bibinfo{person}{Vibhash
  Chandra}, \bibinfo{person}{Kanav Mehra}, \bibinfo{person}{Asit~Kumar Das},
  \bibinfo{person}{Tanmoy Chakraborty}, {and} \bibinfo{person}{Saptarshi
  Ghosh}.} \bibinfo{year}{2018}\natexlab{}.
\newblock \showarticletitle{Ensemble Algorithms for Microblog Summarization}.
\newblock \bibinfo{journal}{\emph{IEEE Intelligent Systems}}
  \bibinfo{volume}{33}, \bibinfo{number}{3} (\bibinfo{year}{2018}),
  \bibinfo{pages}{4--14}.
\newblock


\bibitem[Erkan and Radev(2004)]%
        {erkan2004lexrank}
\bibfield{author}{\bibinfo{person}{G{\"u}nes Erkan} {and}
  \bibinfo{person}{Dragomir~R Radev}.} \bibinfo{year}{2004}\natexlab{}.
\newblock \showarticletitle{Lexrank: Graph-based lexical centrality as salience
  in text summarization}.
\newblock \bibinfo{journal}{\emph{Journal of artificial intelligence research}}
   \bibinfo{volume}{22} (\bibinfo{year}{2004}), \bibinfo{pages}{457--479}.
\newblock


\bibitem[Fabbri et~al\mbox{.}(2021)]%
        {fabbri2021summeval}
\bibfield{author}{\bibinfo{person}{Alexander~R Fabbri},
  \bibinfo{person}{Wojciech Kry{\'s}ci{\'n}ski}, \bibinfo{person}{Bryan
  McCann}, \bibinfo{person}{Caiming Xiong}, \bibinfo{person}{Richard Socher},
  {and} \bibinfo{person}{Dragomir Radev}.} \bibinfo{year}{2021}\natexlab{}.
\newblock \showarticletitle{Summeval: Re-evaluating summarization evaluation}.
\newblock \bibinfo{journal}{\emph{Transactions of the Association for
  Computational Linguistics}}  \bibinfo{volume}{9} (\bibinfo{year}{2021}),
  \bibinfo{pages}{391--409}.
\newblock


\bibitem[Fabbri et~al\mbox{.}(2019)]%
        {fabbri2019multi}
\bibfield{author}{\bibinfo{person}{Alexander~R Fabbri}, \bibinfo{person}{Irene
  Li}, \bibinfo{person}{Tianwei She}, \bibinfo{person}{Suyi Li}, {and}
  \bibinfo{person}{Dragomir~R Radev}.} \bibinfo{year}{2019}\natexlab{}.
\newblock \showarticletitle{Multi-news: A large-scale multi-document
  summarization dataset and abstractive hierarchical model}.
\newblock \bibinfo{journal}{\emph{arXiv preprint arXiv:1906.01749}}
  (\bibinfo{year}{2019}).
\newblock


\bibitem[Gunaratne et~al\mbox{.}(2019)]%
        {gunaratne2019temporal}
\bibfield{author}{\bibinfo{person}{Keith Gunaratne}, \bibinfo{person}{Eric~A
  Coomes}, {and} \bibinfo{person}{Hourmazd Haghbayan}.}
  \bibinfo{year}{2019}\natexlab{}.
\newblock \showarticletitle{Temporal trends in anti-vaccine discourse on
  Twitter}.
\newblock \bibinfo{journal}{\emph{Vaccine}} \bibinfo{volume}{37},
  \bibinfo{number}{35} (\bibinfo{year}{2019}), \bibinfo{pages}{4867--4871}.
\newblock


\bibitem[He et~al\mbox{.}(2020)]%
        {he2020tweetsum}
\bibfield{author}{\bibinfo{person}{Ruifang He}, \bibinfo{person}{Liangliang
  Zhao}, {and} \bibinfo{person}{Huanyu Liu}.} \bibinfo{year}{2020}\natexlab{}.
\newblock \showarticletitle{TWEETSUM: Event oriented social summarization
  dataset}. In \bibinfo{booktitle}{\emph{Proceedings of the 28th International
  Conference on Computational Linguistics}}. \bibinfo{pages}{5731--5736}.
\newblock


\bibitem[Johnson et~al\mbox{.}(2016)]%
        {johnson2016mimic}
\bibfield{author}{\bibinfo{person}{Alistair~EW Johnson}, \bibinfo{person}{Tom~J
  Pollard}, \bibinfo{person}{Lu Shen}, \bibinfo{person}{Li-wei~H Lehman},
  \bibinfo{person}{Mengling Feng}, \bibinfo{person}{Mohammad Ghassemi},
  \bibinfo{person}{Benjamin Moody}, \bibinfo{person}{Peter Szolovits},
  \bibinfo{person}{Leo Anthony~Celi}, {and} \bibinfo{person}{Roger~G Mark}.}
  \bibinfo{year}{2016}\natexlab{}.
\newblock \showarticletitle{MIMIC-III, a freely accessible critical care
  database}.
\newblock \bibinfo{journal}{\emph{Scientific data}} \bibinfo{volume}{3},
  \bibinfo{number}{1} (\bibinfo{year}{2016}), \bibinfo{pages}{1--9}.
\newblock


\bibitem[Johnson et~al\mbox{.}(2020)]%
        {johnson2020online}
\bibfield{author}{\bibinfo{person}{Neil~F Johnson}, \bibinfo{person}{Nicolas
  Vel{\'a}squez}, \bibinfo{person}{Nicholas~Johnson Restrepo},
  \bibinfo{person}{Rhys Leahy}, \bibinfo{person}{Nicholas Gabriel},
  \bibinfo{person}{Sara El~Oud}, \bibinfo{person}{Minzhang Zheng},
  \bibinfo{person}{Pedro Manrique}, \bibinfo{person}{Stefan Wuchty}, {and}
  \bibinfo{person}{Yonatan Lupu}.} \bibinfo{year}{2020}\natexlab{}.
\newblock \showarticletitle{The online competition between pro-and
  anti-vaccination views}.
\newblock \bibinfo{journal}{\emph{Nature}} \bibinfo{volume}{582},
  \bibinfo{number}{7811} (\bibinfo{year}{2020}), \bibinfo{pages}{230--233}.
\newblock


\bibitem[Lewis et~al\mbox{.}(2004)]%
        {lewis2004rcv1}
\bibfield{author}{\bibinfo{person}{David~D Lewis}, \bibinfo{person}{Yiming
  Yang}, \bibinfo{person}{Tony Russell-Rose}, {and} \bibinfo{person}{Fan Li}.}
  \bibinfo{year}{2004}\natexlab{}.
\newblock \showarticletitle{Rcv1: A new benchmark collection for text
  categorization research}.
\newblock \bibinfo{journal}{\emph{Journal of machine learning research}}
  \bibinfo{volume}{5}, \bibinfo{number}{Apr} (\bibinfo{year}{2004}),
  \bibinfo{pages}{361--397}.
\newblock


\bibitem[Lewis et~al\mbox{.}(2020)]%
        {lewis2020bart}
\bibfield{author}{\bibinfo{person}{Mike Lewis}, \bibinfo{person}{Yinhan Liu},
  \bibinfo{person}{Naman Goyal}, \bibinfo{person}{Marjan Ghazvininejad},
  \bibinfo{person}{Abdelrahman Mohamed}, \bibinfo{person}{Omer Levy},
  \bibinfo{person}{Veselin Stoyanov}, {and} \bibinfo{person}{Luke
  Zettlemoyer}.} \bibinfo{year}{2020}\natexlab{}.
\newblock \showarticletitle{BART: Denoising Sequence-to-Sequence Pre-training
  for Natural Language Generation, Translation, and Comprehension}. In
  \bibinfo{booktitle}{\emph{Proceedings of the 58th Annual Meeting of the
  Association for Computational Linguistics}}. \bibinfo{pages}{7871--7880}.
\newblock


\bibitem[Li et~al\mbox{.}(2021)]%
        {li2021heterogeneous}
\bibfield{author}{\bibinfo{person}{Irene Li}, \bibinfo{person}{Tianxiao Li},
  \bibinfo{person}{Yixin Li}, \bibinfo{person}{Ruihai Dong}, {and}
  \bibinfo{person}{Toyotaro Suzumura}.} \bibinfo{year}{2021}\natexlab{}.
\newblock \showarticletitle{Heterogeneous Graph Neural Networks for Multi-label
  Text Classification}.
\newblock \bibinfo{journal}{\emph{arXiv preprint arXiv:2103.14620}}
  (\bibinfo{year}{2021}).
\newblock


\bibitem[Mathew et~al\mbox{.}(2021)]%
        {mathew2021hatexplain}
\bibfield{author}{\bibinfo{person}{Binny Mathew}, \bibinfo{person}{Punyajoy
  Saha}, \bibinfo{person}{Seid~Muhie Yimam}, \bibinfo{person}{Chris Biemann},
  \bibinfo{person}{Pawan Goyal}, {and} \bibinfo{person}{Animesh Mukherjee}.}
  \bibinfo{year}{2021}\natexlab{}.
\newblock \showarticletitle{HateXplain: A Benchmark Dataset for Explainable
  Hate Speech Detection}. In \bibinfo{booktitle}{\emph{Proceedings of the AAAI
  Conference on Artificial Intelligence}}, Vol.~\bibinfo{volume}{35}.
  \bibinfo{pages}{14867--14875}.
\newblock


\bibitem[McCreadie et~al\mbox{.}(2019)]%
        {mccreadie2019trec}
\bibfield{author}{\bibinfo{person}{Richard McCreadie}, \bibinfo{person}{Cody
  Buntain}, {and} \bibinfo{person}{Ian Soboroff}.}
  \bibinfo{year}{2019}\natexlab{}.
\newblock \showarticletitle{Trec incident streams: Finding actionable
  information on social media}.
\newblock  (\bibinfo{year}{2019}).
\newblock


\bibitem[Mitra et~al\mbox{.}(2016)]%
        {mitra2016understanding}
\bibfield{author}{\bibinfo{person}{Tanushree Mitra}, \bibinfo{person}{Scott
  Counts}, {and} \bibinfo{person}{James~W Pennebaker}.}
  \bibinfo{year}{2016}\natexlab{}.
\newblock \showarticletitle{Understanding anti-vaccination attitudes in social
  media}. In \bibinfo{booktitle}{\emph{Tenth International AAAI Conference on
  Web and Social Media}}.
\newblock


\bibitem[Mohammad et~al\mbox{.}(2018)]%
        {mohammad2018semeval}
\bibfield{author}{\bibinfo{person}{Saif Mohammad}, \bibinfo{person}{Felipe
  Bravo-Marquez}, \bibinfo{person}{Mohammad Salameh}, {and}
  \bibinfo{person}{Svetlana Kiritchenko}.} \bibinfo{year}{2018}\natexlab{}.
\newblock \showarticletitle{Semeval-2018 task 1: Affect in tweets}. In
  \bibinfo{booktitle}{\emph{Proceedings of the 12th international workshop on
  semantic evaluation}}. \bibinfo{pages}{1--17}.
\newblock


\bibitem[Mullenbach et~al\mbox{.}(2018)]%
        {mullenbach2018explainable}
\bibfield{author}{\bibinfo{person}{James Mullenbach}, \bibinfo{person}{Sarah
  Wiegreffe}, \bibinfo{person}{Jon Duke}, \bibinfo{person}{Jimeng Sun}, {and}
  \bibinfo{person}{Jacob Eisenstein}.} \bibinfo{year}{2018}\natexlab{}.
\newblock \showarticletitle{Explainable Prediction of Medical Codes from
  Clinical Text}. In \bibinfo{booktitle}{\emph{Proceedings of NAACL-HLT}}.
  \bibinfo{pages}{1101--1111}.
\newblock


\bibitem[M{\"u}ller et~al\mbox{.}(2020)]%
        {muller2020covid}
\bibfield{author}{\bibinfo{person}{Martin M{\"u}ller}, \bibinfo{person}{Marcel
  Salath{\'e}}, {and} \bibinfo{person}{Per~E Kummervold}.}
  \bibinfo{year}{2020}\natexlab{}.
\newblock \showarticletitle{Covid-twitter-bert: A natural language processing
  model to analyse covid-19 content on twitter}.
\newblock \bibinfo{journal}{\emph{arXiv preprint arXiv:2005.07503}}
  (\bibinfo{year}{2020}).
\newblock


\bibitem[M{\"u}ller and Salath{\'e}(2019)]%
        {muller2019crowdbreaks}
\bibfield{author}{\bibinfo{person}{Martin~M M{\"u}ller} {and}
  \bibinfo{person}{Marcel Salath{\'e}}.} \bibinfo{year}{2019}\natexlab{}.
\newblock \showarticletitle{Crowdbreaks: tracking health trends using public
  social media data and crowdsourcing}.
\newblock \bibinfo{journal}{\emph{Frontiers in public health}}
  \bibinfo{volume}{7} (\bibinfo{year}{2019}), \bibinfo{pages}{81}.
\newblock


\bibitem[Nguyen et~al\mbox{.}(2018)]%
        {nguyen2018tsix}
\bibfield{author}{\bibinfo{person}{Minh-Tien Nguyen}, \bibinfo{person}{Dac~Viet
  Lai}, \bibinfo{person}{Huy~Tien Nguyen}, {and} \bibinfo{person}{Minh
  Le~Nguyen}.} \bibinfo{year}{2018}\natexlab{}.
\newblock \showarticletitle{Tsix: a human-involved-creation dataset for tweet
  summarization}. In \bibinfo{booktitle}{\emph{Proc. International Conference
  on Language Resources and Evaluation (LREC)}}.
\newblock


\bibitem[Nuzhath et~al\mbox{.}(2020)]%
        {nuzhath2020covid}
\bibfield{author}{\bibinfo{person}{Tasmiah Nuzhath}, \bibinfo{person}{Samia
  Tasnim}, \bibinfo{person}{Rahul~Kumar Sanjwal},
  \bibinfo{person}{Nusrat~Fahmida Trisha}, \bibinfo{person}{Mariya Rahman},
  \bibinfo{person}{SM~Farabi Mahmud}, \bibinfo{person}{Arif Arman},
  \bibinfo{person}{Susmita Chakraborty}, {and} \bibinfo{person}{Md~Mahbub
  Hossain}.} \bibinfo{year}{2020}\natexlab{}.
\newblock \showarticletitle{COVID-19 vaccination hesitancy, misinformation and
  conspiracy theories on social media: A content analysis of Twitter data}.
\newblock  (\bibinfo{year}{2020}).
\newblock


\bibitem[Paul et~al\mbox{.}(2021)]%
        {paul2021attitudes}
\bibfield{author}{\bibinfo{person}{Elise Paul}, \bibinfo{person}{Andrew
  Steptoe}, {and} \bibinfo{person}{Daisy Fancourt}.}
  \bibinfo{year}{2021}\natexlab{}.
\newblock \showarticletitle{Attitudes towards vaccines and intention to
  vaccinate against COVID-19: Implications for public health communications}.
\newblock \bibinfo{journal}{\emph{The Lancet Regional Health-Europe}}
  \bibinfo{volume}{1} (\bibinfo{year}{2021}), \bibinfo{pages}{100012}.
\newblock


\bibitem[Pedregosa et~al\mbox{.}(2011)]%
        {scikit-learn}
\bibfield{author}{\bibinfo{person}{F. Pedregosa}, \bibinfo{person}{G.
  Varoquaux}, \bibinfo{person}{A. Gramfort}, \bibinfo{person}{V. Michel},
  \bibinfo{person}{B. Thirion}, \bibinfo{person}{O. Grisel},
  \bibinfo{person}{M. Blondel}, \bibinfo{person}{P. Prettenhofer},
  \bibinfo{person}{R. Weiss}, \bibinfo{person}{V. Dubourg}, \bibinfo{person}{J.
  Vanderplas}, \bibinfo{person}{A. Passos}, \bibinfo{person}{D. Cournapeau},
  \bibinfo{person}{M. Brucher}, \bibinfo{person}{M. Perrot}, {and}
  \bibinfo{person}{E. Duchesnay}.} \bibinfo{year}{2011}\natexlab{}.
\newblock \showarticletitle{Scikit-learn: Machine Learning in {P}ython}.
\newblock \bibinfo{journal}{\emph{Journal of Machine Learning Research}}
  \bibinfo{volume}{12} (\bibinfo{year}{2011}), \bibinfo{pages}{2825--2830}.
\newblock


\bibitem[Poddar et~al\mbox{.}(2022)]%
        {poddar2022winds}
\bibfield{author}{\bibinfo{person}{Soham Poddar}, \bibinfo{person}{Mainack
  Mondal}, \bibinfo{person}{Janardan Misra}, \bibinfo{person}{Niloy Ganguly},
  {and} \bibinfo{person}{Saptarshi Ghosh}.} \bibinfo{year}{2022}\natexlab{}.
\newblock \showarticletitle{Winds of Change: Impact of COVID-19 on
  Vaccine-related Opinions of Twitter users}. In
  \bibinfo{booktitle}{\emph{Proceedings of the Sixteenth International AAAI
  Conference on Web and Social Media (ICWSM'22)}}.
\newblock


\bibitem[Praveen et~al\mbox{.}(2021)]%
        {praveen2021analyzing}
\bibfield{author}{\bibinfo{person}{SV Praveen}, \bibinfo{person}{Rajesh
  Ittamalla}, {and} \bibinfo{person}{Gerard Deepak}.}
  \bibinfo{year}{2021}\natexlab{}.
\newblock \showarticletitle{Analyzing the attitude of Indian citizens towards
  COVID-19 vaccine--A text analytics study}.
\newblock \bibinfo{journal}{\emph{Diabetes \& Metabolic Syndrome: Clinical
  Research \& Reviews}} \bibinfo{volume}{15}, \bibinfo{number}{2}
  (\bibinfo{year}{2021}), \bibinfo{pages}{595--599}.
\newblock


\bibitem[Raffel et~al\mbox{.}(2019)]%
        {raffel2019exploring}
\bibfield{author}{\bibinfo{person}{Colin Raffel}, \bibinfo{person}{Noam
  Shazeer}, \bibinfo{person}{Adam Roberts}, \bibinfo{person}{Katherine Lee},
  \bibinfo{person}{Sharan Narang}, \bibinfo{person}{Michael Matena},
  \bibinfo{person}{Yanqi Zhou}, \bibinfo{person}{Wei Li}, {and}
  \bibinfo{person}{Peter~J Liu}.} \bibinfo{year}{2019}\natexlab{}.
\newblock \showarticletitle{Exploring the limits of transfer learning with a
  unified text-to-text transformer}.
\newblock \bibinfo{journal}{\emph{arXiv preprint arXiv:1910.10683}}
  (\bibinfo{year}{2019}).
\newblock


\bibitem[Ribeiro et~al\mbox{.}(2016)]%
        {ribeiro2016should}
\bibfield{author}{\bibinfo{person}{Marco~Tulio Ribeiro},
  \bibinfo{person}{Sameer Singh}, {and} \bibinfo{person}{Carlos Guestrin}.}
  \bibinfo{year}{2016}\natexlab{}.
\newblock \showarticletitle{``Why should i trust you?'' Explaining the
  predictions of any classifier}. In \bibinfo{booktitle}{\emph{Proc. ACM SIGKDD
  Conference on Knowledge Discovery and Data Mining}}.
\newblock


\bibitem[Rudra et~al\mbox{.}(2015)]%
        {rudra2015extracting}
\bibfield{author}{\bibinfo{person}{Koustav Rudra}, \bibinfo{person}{Subham
  Ghosh}, \bibinfo{person}{Niloy Ganguly}, \bibinfo{person}{Pawan Goyal}, {and}
  \bibinfo{person}{Saptarshi Ghosh}.} \bibinfo{year}{2015}\natexlab{}.
\newblock \showarticletitle{Extracting situational information from microblogs
  during disaster events: a classification-summarization approach}. In
  \bibinfo{booktitle}{\emph{Proceedings of the 24th ACM international on
  conference on information and knowledge management}}.
  \bibinfo{pages}{583--592}.
\newblock


\bibitem[Sechidis et~al\mbox{.}(2011)]%
        {sechidis2011stratification}
\bibfield{author}{\bibinfo{person}{Konstantinos Sechidis},
  \bibinfo{person}{Grigorios Tsoumakas}, {and} \bibinfo{person}{Ioannis
  Vlahavas}.} \bibinfo{year}{2011}\natexlab{}.
\newblock \showarticletitle{On the stratification of multi-label data}. In
  \bibinfo{booktitle}{\emph{Joint European Conference on Machine Learning and
  Knowledge Discovery in Databases}}. Springer, \bibinfo{pages}{145--158}.
\newblock


\bibitem[Shahsavari et~al\mbox{.}(2020)]%
        {shahsavari2020conspiracy}
\bibfield{author}{\bibinfo{person}{Shadi Shahsavari}, \bibinfo{person}{Pavan
  Holur}, \bibinfo{person}{Tianyi Wang}, \bibinfo{person}{Timothy~R
  Tangherlini}, {and} \bibinfo{person}{Vwani Roychowdhury}.}
  \bibinfo{year}{2020}\natexlab{}.
\newblock \showarticletitle{Conspiracy in the time of corona: Automatic
  detection of emerging COVID-19 conspiracy theories in social media and the
  news}.
\newblock \bibinfo{journal}{\emph{Journal of computational social science}}
  \bibinfo{volume}{3}, \bibinfo{number}{2} (\bibinfo{year}{2020}),
  \bibinfo{pages}{279--317}.
\newblock


\bibitem[Sonawane et~al\mbox{.}(2021)]%
        {sonawane2021covid}
\bibfield{author}{\bibinfo{person}{Kalyani Sonawane},
  \bibinfo{person}{Catherine~L Troisi}, {and} \bibinfo{person}{Ashish~A
  Deshmukh}.} \bibinfo{year}{2021}\natexlab{}.
\newblock \showarticletitle{COVID-19 vaccination in the UK: Addressing vaccine
  hesitancy}.
\newblock \bibinfo{journal}{\emph{The Lancet Regional Health--Europe}}
  \bibinfo{volume}{1} (\bibinfo{year}{2021}).
\newblock


\bibitem[Staff(2021a)]%
        {ajmc2020timeline}
\bibfield{author}{\bibinfo{person}{AJMC Staff}.}
  \bibinfo{year}{2021}\natexlab{a}.
\newblock \showarticletitle{A Timeline of COVID-19 Developments in 2020}.
\newblock \bibinfo{journal}{\emph{AJMC}} (\bibinfo{year}{2021}).
\newblock
\urldef\tempurl%
\url{https://www.ajmc.com/view/a-timeline-of-covid19-developments-in-2020}
\showURL{%
\tempurl}


\bibitem[Staff(2021b)]%
        {ajmc2021timeline}
\bibfield{author}{\bibinfo{person}{AJMC Staff}.}
  \bibinfo{year}{2021}\natexlab{b}.
\newblock \showarticletitle{A Timeline of COVID-19 Vaccine Developments in
  2021}.
\newblock \bibinfo{journal}{\emph{AJMC}} (\bibinfo{year}{2021}).
\newblock
\urldef\tempurl%
\url{https://www.ajmc.com/view/a-timeline-of-covid-19-vaccine-developments-in-2021}
\showURL{%
\tempurl}


\bibitem[Szyma{\'n}ski and Kajdanowicz(2017a)]%
        {szymanski2017network}
\bibfield{author}{\bibinfo{person}{Piotr Szyma{\'n}ski} {and}
  \bibinfo{person}{Tomasz Kajdanowicz}.} \bibinfo{year}{2017}\natexlab{a}.
\newblock \showarticletitle{A network perspective on stratification of
  multi-label data}. In \bibinfo{booktitle}{\emph{First International Workshop
  on Learning with Imbalanced Domains: Theory and Applications}}. PMLR,
  \bibinfo{pages}{22--35}.
\newblock


\bibitem[Szyma{\'n}ski and Kajdanowicz(2017b)]%
        {szymanski2017scikit}
\bibfield{author}{\bibinfo{person}{Piotr Szyma{\'n}ski} {and}
  \bibinfo{person}{Tomasz Kajdanowicz}.} \bibinfo{year}{2017}\natexlab{b}.
\newblock \showarticletitle{A scikit-based Python environment for performing
  multi-label classification}.
\newblock \bibinfo{journal}{\emph{arXiv preprint arXiv:1702.01460}}
  (\bibinfo{year}{2017}).
\newblock


\bibitem[Tao et~al\mbox{.}(2013)]%
        {tao2013groundhog}
\bibfield{author}{\bibinfo{person}{Ke Tao}, \bibinfo{person}{Fabian Abel},
  \bibinfo{person}{Claudia Hauff}, \bibinfo{person}{Geert-Jan Houben}, {and}
  \bibinfo{person}{Ujwal Gadiraju}.} \bibinfo{year}{2013}\natexlab{}.
\newblock \showarticletitle{Groundhog day: near-duplicate detection on
  twitter}. In \bibinfo{booktitle}{\emph{Proceedings of the 22nd international
  conference on World Wide Web}}. \bibinfo{pages}{1273--1284}.
\newblock


\bibitem[Troiano and Nardi(2021)]%
        {troiano2021vaccine}
\bibfield{author}{\bibinfo{person}{Gianmarco Troiano} {and}
  \bibinfo{person}{Alessandra Nardi}.} \bibinfo{year}{2021}\natexlab{}.
\newblock \showarticletitle{Vaccine hesitancy in the era of COVID-19}.
\newblock \bibinfo{journal}{\emph{Public health}}  \bibinfo{volume}{194}
  (\bibinfo{year}{2021}), \bibinfo{pages}{245--251}.
\newblock


\bibitem[Wang et~al\mbox{.}(2021)]%
        {Wang_Gao_Bai_Lapata_Huang_2021}
\bibfield{author}{\bibinfo{person}{Haonan Wang}, \bibinfo{person}{Yang Gao},
  \bibinfo{person}{Yu Bai}, \bibinfo{person}{Mirella Lapata}, {and}
  \bibinfo{person}{Heyan Huang}.} \bibinfo{year}{2021}\natexlab{}.
\newblock \showarticletitle{Exploring Explainable Selection to Control
  Abstractive Summarization}. In \bibinfo{booktitle}{\emph{Proc. AAAI
  Conference on Artificial Intelligence}}. \bibinfo{pages}{13933--13941}.
\newblock


\bibitem[Yuan et~al\mbox{.}(2019)]%
        {yuan2019examining}
\bibfield{author}{\bibinfo{person}{Xiaoyi Yuan}, \bibinfo{person}{Ross~J
  Schuchard}, {and} \bibinfo{person}{Andrew~T Crooks}.}
  \bibinfo{year}{2019}\natexlab{}.
\newblock \showarticletitle{Examining emergent communities and social bots
  within the polarized online vaccination debate in Twitter}.
\newblock \bibinfo{journal}{\emph{Social media+ society}} \bibinfo{volume}{5},
  \bibinfo{number}{3} (\bibinfo{year}{2019}),
  \bibinfo{pages}{2056305119865465}.
\newblock


\bibitem[Zhang et~al\mbox{.}(2020)]%
        {zhang2020pegasus}
\bibfield{author}{\bibinfo{person}{Jingqing Zhang}, \bibinfo{person}{Yao Zhao},
  \bibinfo{person}{Mohammad Saleh}, {and} \bibinfo{person}{Peter Liu}.}
  \bibinfo{year}{2020}\natexlab{}.
\newblock \showarticletitle{Pegasus: Pre-training with extracted gap-sentences
  for abstractive summarization}. In \bibinfo{booktitle}{\emph{International
  Conference on Machine Learning}}. PMLR, \bibinfo{pages}{11328--11339}.
\newblock


\bibitem[Zhang et~al\mbox{.}(2021)]%
        {zhang2021explain}
\bibfield{author}{\bibinfo{person}{Zijian Zhang}, \bibinfo{person}{Koustav
  Rudra}, {and} \bibinfo{person}{Avishek Anand}.}
  \bibinfo{year}{2021}\natexlab{}.
\newblock \showarticletitle{Explain and predict, and then predict again}. In
  \bibinfo{booktitle}{\emph{Proceedings of the 14th ACM International
  Conference on Web Search and Data Mining}}. \bibinfo{pages}{418--426}.
\newblock


\bibitem[Zheng and Lapata(2019)]%
        {zheng2019sentence}
\bibfield{author}{\bibinfo{person}{Hao Zheng} {and} \bibinfo{person}{Mirella
  Lapata}.} \bibinfo{year}{2019}\natexlab{}.
\newblock \showarticletitle{Sentence Centrality Revisited for Unsupervised
  Summarization}. In \bibinfo{booktitle}{\emph{Proceedings of the 57th Annual
  Meeting of the Association for Computational Linguistics}}.
  \bibinfo{pages}{6236--6247}.
\newblock


\end{thebibliography}

%%
%% If your work has an appendix, this is the place to put it.
% \appendix

\end{document}